\documentclass{article}
\usepackage{arxiv}

\usepackage[utf8]{inputenc}
\usepackage[T1]{fontenc}
\usepackage{hyperref}
\usepackage{url}
\usepackage{booktabs}
\usepackage{amsfonts}
\usepackage{nicefrac}
\usepackage{microtype}
\usepackage{graphicx}
\usepackage{natbib}
\usepackage{doi}
\usepackage{multirow}
\usepackage{amsmath}
\usepackage{cleveref}
\usepackage{array}
\graphicspath{{figures/}}

\title{Prompt, Probe, Train, or Annotate? \\ \large Single-camera sports video understanding in amateur settings}

\date{}

\author{
	\href{https://orcid.org/0009-0002-3124-633X}{\includegraphics[scale=0.06]{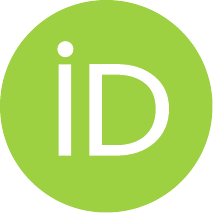}\hspace{1mm}Sai Varun Kodathala} \\
	Research and Development\\
	Sports Vision, Inc.\\
	Minnetonka, MN \\
	\texttt{varun@sportsvision.ai} \\
	\And
	Prashanth Pollishetty\\
	Research and Development\\
	Sports Vision, Inc.\\
	Minnetonka, MN \\
	\texttt{prashanth@sportsvision.ai} \\
	\And
	Jaylen Cargill\\
	Research and Development\\
	Sports Vision, Inc.\\
	Minnetonka, MN \\
	\texttt{Jaylen@sportsvision.ai} \\
}

\renewcommand{\headeright}{}
\renewcommand{\undertitle}{}
\renewcommand{\shorttitle}{}

\hypersetup{
	pdftitle={Prompt, Probe, Train, or Annotate?},
	pdfsubject={cs.CV, cs.LG},
	pdfauthor={Sai Varun Kodathala, Prashanth Pollishetty, Jaylen Cargill},
	pdfkeywords={Video Understanding, Action Recognition, Player Identification, Vision-Language Models,
	World Models, Human-in-the-Loop, Production Machine Learning},
}

\begin{document}
\maketitle

\begin{abstract}
Video understanding is usually benchmarked on curated, single-actor, or professionally filmed
clips, and a model's strong score on those benchmarks is routinely read as evidence that it is
robust enough for deployment. Amateur team sport is a useful, and largely untested, place to check
that assumption: over eight million students played a school sport in the United States in
2024--25 alone, almost none of it captured by anything more than a single fixed camera a coach or
parent happened to point at the court, with several candidate actors crowded into frame and no
operator, replay, or second angle to fall back on. We use this setting, and volleyball specifically,
to ask a question that generalises well beyond one sport: does strong performance on general video
and world-model benchmarks translate into reliable, per-actor attribution once the footage is this
chaotic. Turning such footage into per-player statistics is not one task but a chain of them:
finding the boundaries between plays, spotting each contact, classifying what happened, deciding
the outcome, assigning the team, naming the individual, and totalling the result without letting
errors cancel each other out. We evaluate four ways of doing this (video understanding through prompting and agentic
reasoning over frontier vision-language models, composing classical computer-vision components with small trained
specialists, probing self-supervised video world models, and manually annotating the footage) at
every stage of that chain, on the same corpus and under the same cost accounting. The corpus is 66
amateur volleyball matches with 46{,}648 human-labelled contacts, filmed under conditions no
published sports-video benchmark uses and no commercially available system, on the evidence of its
own documentation, actually operates without a person closing the gap somewhere between camera and
box score. No single paradigm is best at every stage, and static, single-frame computer vision on
its own is not competitive at any stage that involves motion or identity. A prompted model segments
matches into plays about as well as anything we tested, yet a two-million-parameter trained model
outperforms it at spotting individual contacts, at a small fraction of the cost. Outcomes cannot be
read from the pixels at all (the ball regularly measures under thirty pixels across), but the
game's own rules recover the winner of a rally correctly essentially every time, because volleyball
fixes who serves next once a rally ends. Identity is where every automated approach struggles: a
fine-tuned vision-language model converges to a low, flat accuracy and almost never declines to
answer even when it is guessing; a frozen video world-model probe, despite strong scores on general
action and prediction benchmarks, barely outperforms a baseline built from nothing but bounding-box
coordinates once identity rather than action is what is being asked; and nine commercial readers
spanning two orders of magnitude in price differ by only about ten points of accuracy, with the
most expensive reader not the most accurate one. We trace the identity result to a distinction we
believe is under-appreciated in sports video specifically: an athletic action is a repeated, trained
motor pattern whose temporal signature a model can exploit, while a jersey number is a static fact a
temporal signal cannot recover if it was never legibly visible to begin with, which is why holistic,
context-rich reasoning measurably improves event detection while leaving identity essentially
unchanged, and why benchmark robustness on one kind of video task does not imply robustness on the
other. Manual annotation, examined here as a fourth option rather than assumed as free, instant
ground truth, carries its own quantifiable cost and latency. We close with a stage-by-stage account of where each
approach is worth its cost, an evaluation protocol built to expose exactly these differences rather
than average over them, and a discussion of what does and does not transfer to sports video
recorded outside a broadcast studio, and beyond volleyball to amateur sport more generally.

\end{abstract}


\section{Introduction}
\label{sec:intro}

Sport is one of the largest organised activities young people take part in, and almost none of it
is played in front of a camera anyone is paid to operate. Figure~\ref{fig:context}, drawn directly
from our own corpus, shows what that actually looks like: a single fixed camera, several candidate
actors crowded into frame at once, and no operator, replay, or second angle to fall back on. In the United States alone, over eight
million students competed in a school sport during the 2024--25 academic year, an all-time high
\citep{nfhs2025participation}. Volleyball is a sizeable part of that figure on its own: nearly half a
million girls and close to a hundred thousand boys played it at the high school level in that same
year \citep{nfhs2025participation}, before any youth club, recreational league, or college
intramural program is counted at all. These programs are not, for the most part, well
resourced. Athletic departments at even successful, well-run schools typically allocate only two to
four percent of a school's total budget to sport \citep{aspen2024schoolsports}, and the gap between
that allocation and what a season actually costs is routinely closed by booster-club fundraising,
gate receipts from the one or two sports that draw a crowd, and per-season fees of several hundred
dollars passed directly to families \citep{aspen2024schoolsports}. Fewer than four in ten students
at a public high school play a sport at all, and the number of schools offering one has, in some
recent years, fallen even as the broader economy grew \citep{aspen2024schoolsports}. This is the
setting the overwhelming majority of organised sport actually happens in, and it is a setting a
booster club's annual budget cannot stretch to cover with an installed multi-camera rig, a
professional camera operator, or a subscription to hand-tagged analyst review.

That scarcity is also unevenly distributed, in ways a national average obscures. Within a single
state, district-level athletics spending has been documented to vary by a factor of roughly two to
three, with the highest-spending public high schools in one state analysis outspending the
lowest by close to three times per athlete, and a participation gap that tracks it: schools serving
mostly higher-income communities have been documented fielding sport-season participation rates more
than double those of the state's lowest-income schools \citep{bellwether2025unequalfields}. The
people asked to close the resulting gap between what a season needs and what a program can pay are
themselves paid for only a fraction of the time the job requires. A state school-boards association's
own guidance to district administrators works the arithmetic directly: a typical coaching stipend
clears the state minimum wage only up to roughly three hundred hours of coaching time across a
season, a figure a full slate of practices, matches, travel, and film review can exceed well before
the season ends, after which the coach is, in effect, working below minimum wage for the remainder
of it \citep{wasb2022stipends}. Even the conditions under which the sport is played are not
standardised: two neighbouring states' own athletic-association bylaws set maximum varsity volleyball
match counts roughly thirty-five percent apart from each other in the same season
\citep{nmaa2024volleyball,mshsl2023handbook}, which is one concrete measure of how little about
amateur sport, budget, coaching time, or even the length of a season, can be assumed to generalise
from one program to the next.

Analysis of that same sport, meanwhile, has become a genuine industry when the setting is
professional or broadcast competition. Independent market analysts value the global sports
analytics market in the single-digit billions of dollars as of the mid-2020s and project it to
roughly triple by the end of the decade \citep{grandview2025sportsanalytics}, a figure driven almost
entirely by professional leagues, broadcasters, and the betting and media operations built around
them. The statistics a professional team's analytics department produces are not a luxury; they
shape who plays, who is recruited, and how a season is coached, and there is no reason to think a
high-school or club athlete's development would benefit any less from the same kind of information,
were it available to them at a cost their program could actually pay. It generally is not.
Commercial systems that do offer automated statistics to amateur programs (examined in detail in
Section~\ref{ssec:vendors}) turn out, on inspection of their own documentation, to close the gap
between an ordinary camera and a finished box score with a person: an analyst tagging film by hand,
a parent typing a roster before kickoff, or a coach correcting a queue of uncertain attributions
after each match. This is a reasonable way to ship a product a real program can use today, and we
do not present it as a failure of engineering. It is, however, evidence that the underlying
perception problem (turning an ordinary, single, amateur camera into a reliable account of who did
what) has not yet been solved by computation alone, and this paper measures, in detail, exactly
where that is and is not true.

\begin{figure*}[t]
\centering
\begin{minipage}{0.48\textwidth}
\centering
\includegraphics[width=\textwidth]{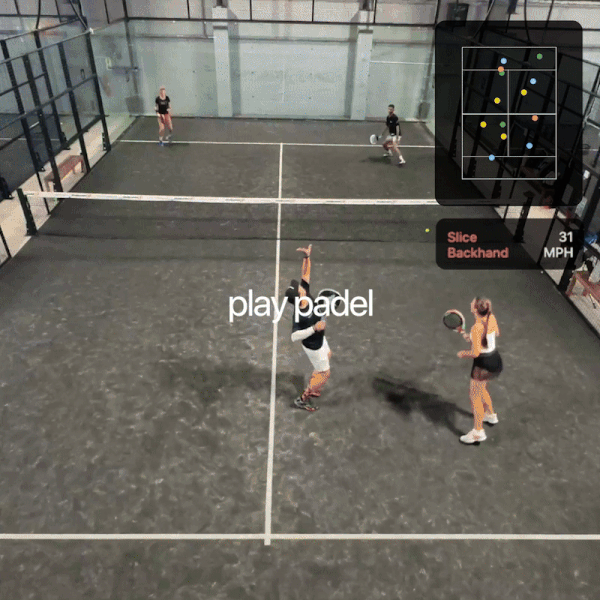}\\[2pt]
{\footnotesize (a) Two-actor court, a shot attached to a stat}
\end{minipage}\hfill
\begin{minipage}{0.48\textwidth}
\centering
\includegraphics[width=\textwidth]{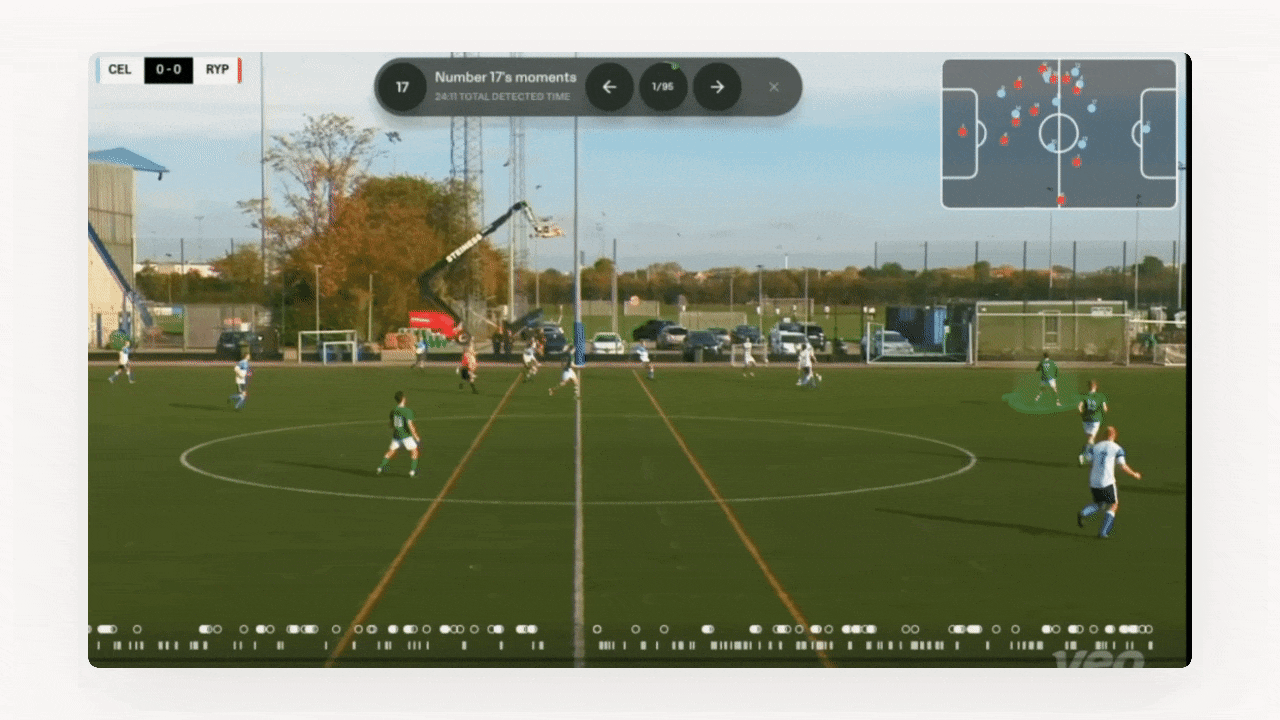}\\[2pt]
{\footnotesize (b) Team-sport frame, detection with no stat}
\end{minipage}
\caption{Two vendors' own published interfaces, reproduced here for commentary, at opposite ends of
the actor-count progression described in Section~\ref{sec:intro}. (a) A paired-opponent racket
sport, where the net fixes which half of the court each player occupies: a detected shot is
directly labelled with a specific statistic, ``Slice Backhand, 31 MPH'' \citep{swingvision2026site}.
(b) A team-sport frame from a different vendor: the system has correctly detected and followed
player number 17 for 24 minutes and 11 seconds of match time and plots every detected player's
position as a coloured dot on a schematic pitch, but neither this screen nor any screen it links to
attaches a single per-player statistic, kills, passes, distance, anything with a number attached,
to that detection \citep{veo2026spotlight}. The two screenshots illustrate, in the market's own
published interfaces rather than in our claim about them, exactly the divide this paper measures:
detecting and following a player is a different, and in our own results a substantially easier,
problem than knowing what that player did, and the divide opens as soon as a court geometry no
longer fixes identity for free. Accessed 2026-09-23; neither product is otherwise discussed or
compared against in this paper.}
\label{fig:vendordetection}
\end{figure*}

Start with a single golfer on a driving range. One camera, one body, one motion that begins and
ends within about a second and repeats identically from swing to swing. A single-camera golf-swing
benchmark built on exactly this setting reports detecting the eight key events of a swing at better
than three-quarters accuracy from ordinary video alone \citep{mcnally2019golfdb}, and commercial
systems built for the same setting report launch and swing metrics that professional coaches treat
as reliable. The reason is structural rather than algorithmic: there is nothing to attribute,
because there is nobody else the motion could belong to. Move to a tennis court and the problem
gains one dimension (two bodies, one ball, a net that fixes which half of the frame each player
occupies at any moment), and consumer systems again publish accuracy figures with some confidence,
because assigning an event to ``the player on the near side'' or ``the player on the far side'' is
a geometric fact, not a perceptual guess (Figure~\ref{fig:vendordetection}a shows this directly, a
real product attaching a named shot type and a speed to a single detected stroke on exactly this
kind of court). Move again, to a beach volleyball court with two players per side, and the guarantee
weakens: within a team of two, the camera must now distinguish between
two people wearing similar or identical attire, and published accuracy claims all but disappear.
Move once more, to an indoor volleyball court with six players a side, or a basketball court with
five, and no vendor we could find publishes an accuracy number for per-player statistics at all.
Continue to an eleven-a-side football pitch, filmed from a fixed camera with no operator to follow
the play, and the same silence holds (Figure~\ref{fig:vendordetection}b shows exactly this stage: a
different real product correctly detecting and following an individual player for over twenty
minutes of match time, with no per-player statistic attached anywhere on that screen or the pages
it links to), now compounded by a much larger playing surface and a ball
that can vanish behind a cluster of bodies for entire phases of the game. The pattern across this
progression is not that computer vision gets somewhat harder as more people enter the frame. It is
that a specific capability (knowing \emph{which} person did something, as opposed to knowing that
something happened) degrades sharply and then never recovers, and every commercial claim of
automatic per-player statistics we could locate turns out, on inspection of the vendor's own
documentation, to depend on a human being closing that gap: a scorekeeper tapping a name, an
analyst tagging a clip, a parent typing a roster before the match begins.

We arrived at this progression the slow way, through our own earlier work, which explains why the
paper is organised the way it is. Our group's first attempt at
this general problem was on amateur and professional boxing, working with USA Boxing footage under
the US Olympic and Paralympic Committee, and it sat at the two-actor point on the axis above: given
a bout, classify which type of punch had just been thrown. That early system was, in the terms this
paper uses, purely a computer-vision approach at a single stage, a convolutional network trained to
recognise a punch from its visual appearance across a short clip \citep{kodathala2025sixsigma}, and
it worked, in the narrow sense of reaching high accuracy on held-out punches, precisely because a
two-person combat sport removes the identity problem in the same way tennis does: there is one
opponent to attribute an action to, not a roster of six or eleven. What that project made clear,
well before we had the corpus this paper is built on, was that the harder and more valuable problem
sat one step further along the same axis, in sports where a team, not an opponent, occupied each
half of the frame, and where no vendor was willing to publish an accuracy number for exactly the
reason this paper measures directly. We moved next to baseball, an internal effort rather than a
published one, and only then to basketball and volleyball, the setting SV3.3B and the present paper
address. Across that progression, our own models changed shape in a direction this paper's results
justify in hindsight: from a single classifier reasoning over one clip in isolation, toward the
track-level, temporally integrated composition of small models described in
Section~\ref{sec:paradigms}, because it was only once we were working in a full-roster team sport
that treating identity as a separate, harder problem from action recognition, rather than folding
both into one end-to-end model, stopped being an architectural preference and became a necessity.

This paper measures that gap directly, rather than inferring it from vendor silence. We do so on
volleyball because it sits at an instructive point in the progression just described (six players
per side, one ball, a rulebook that constrains who is legally allowed to touch that ball and in
what order), and because we were able to assemble something the field otherwise lacks: 66 amateur
matches with every one of 46{,}648 contacts hand-labelled by a person watching the film, filmed on
a single fixed camera of the kind a parent, an assistant coach, or a school's own gymnasium rig
would actually use. This is deliberately not broadcast footage. It has no operator tracking the
ball, no second camera for a reverse angle, no on-screen graphics naming the server, and none of
the incidental cues (replays, commentary, an official play-by-play feed running in
parallel) that make professional sports video easier to interpret than its resolution alone would
suggest. It is, in other words, the footage that actually exists for the millions of athletes
counted in the participation figures above, and it is almost entirely absent from the datasets the
field uses to claim progress.

This progression reflects a strategic position that is easy to state implicitly and harder to
defend once stated plainly, so we state it plainly here. General intelligence, applied uniformly across
every domain a camera might point at, is not, in our judgement, a viable target for a company at our
scale, and we suspect it is a harder target than current marketing around it admits; even large,
well-resourced efforts to bring sweeping ``intelligence'' branding to a single sport
\citep{ibm2025agassi} have so far shipped narrow, task-specific capability rather than anything
resembling general competence across a sport's full range of demands. Our own bet is the opposite
one: that progress comes from choosing a vertical and committing to its specific structure, rather
than from a model built to generalise everywhere at once, and that sport is an unusually good
vertical to make that bet in. Competitive sport generates an enormous, naturally-occurring volume of
structured behavioural data as a byproduct of simply being played, scores, rosters, rule-constrained
outcomes, recorded matches, at a scale few other human activities produce without deliberate
instrumentation, and the participation figures in this section (Section~\ref{sec:intro}) are one
measure of how large that volume already is before a single frame of video is analysed. If a
domain-specific approach is going to outperform a general one anywhere, a domain this data-rich, this
structurally constrained by its own rules, and this systematically under-served by existing
tools is a reasonable place to expect it, and this paper is our attempt to measure whether
that expectation holds rather than assert that it does.

Given this footage, we ask a narrower question than ``can artificial intelligence understand
sports video,'' because that question has already produced a decade of benchmark scores without
producing a single vendor willing to publish a per-player accuracy figure. We ask instead: at each
distinct stage between raw video and a finished box score (segmenting the match into individual
plays, spotting the contacts within each play, classifying what each contact was, determining how
the play ended, working out which team is which, naming the individual who acted, and totalling
the result), which of four available approaches actually earns its cost? The four are prompting a
frontier vision-language model and letting it reason over the footage directly; composing
classical detection and tracking with small models trained specifically for this footage; probing
a self-supervised video world model, of the kind currently described as a promising route toward
general video understanding; and paying a person to watch the film and write down what
happened, which is what the market does today and which we treat here as a measurable method
rather than as ground truth to be assumed correct. Elsewhere, one of us has already built and
evaluated a lightweight model that generates fluent, technically detailed descriptions of sporting
action from video \citep{vutukoori2025sv33b}; that model, like every other approach we test here,
can say that a spike occurred without being able to say whose spike it was, and the distance
between those two capabilities is the subject of this paper.

Our answer is that no single paradigm is best at every stage, and that the differences between
stages are large enough to change which approach a practitioner should choose depending on what
they are trying to measure. A frontier model segments a match into individual rallies about as
well as any method we tested; the same model, or one much like it, is measurably worse than a
model orders of magnitude smaller at the narrower job of spotting an individual contact within a
rally, and costs orders of magnitude more per match to run. The outcome of a rally (who won
it) cannot be read from the pixels with any reliability at all, because the ball is small enough
and the rally short enough that no method we tried can consistently see where it lands; and yet the
outcome can be recovered almost perfectly from the rules of the sport alone, because volleyball
fixes, as a matter of law rather than observation, which team serves next. Identity is where every
automated method we tested falls furthest short, and we spend the largest part of this paper
explaining exactly why, in terms that generalise well beyond one sport: the information a jersey
number carries is not, at the resolutions an amateur camera actually produces, present in the
pixels in the way a benchmark trained on broadcast close-ups assumes it to be, and, as
Section~\ref{ssec:musclememory} argues at length, identity does not share the temporal, motion-based
structure that makes sporting \emph{action} itself so tractable to recognise. A serve, a dig, or a
spike is a repeated, trained motor pattern (muscle memory, in the athlete's own vocabulary), and
that repetition is exactly the redundancy a temporal model can exploit; a jersey number is a static
fact about a garment that no amount of motion or temporal context can conjure into existence if it
was never legibly visible in the first place.

The paper's structure follows this argument. Section~\ref{sec:related} places the work against
prior sports-video research and against the public claims of the commercial systems already
operating in this space. Section~\ref{sec:setting} describes the corpus and the perceptual
conditions it was filmed under. Section~\ref{sec:stages} decomposes the video-to-statistics task
into its constituent stages, which is the frame the rest of the paper is organised around.
Section~\ref{sec:paradigms} describes the four approaches without disclosing the specific
configuration of the system we operate commercially. Section~\ref{sec:results} reports accuracy
and cost at every stage, including where whole-match reasoning genuinely helps and where it does
not. Section~\ref{sec:failure} explains, with citation to the broader literature on visual
grounding and multi-step inference, why the failures take the shape they do.
Section~\ref{sec:deployment} reports what it costs to run this at the scale of a real product.
Section~\ref{sec:discussion} discusses what transfers to sports beyond volleyball, and what does
not. We close in Section~\ref{sec:limitations} with the limitations of the study, and in the
statements that follow it with what a paper built on private, identifiable footage of youth
athletes owes its readers.

\section{Related Work}
\label{sec:related}

\subsection{From one body to a field of them}

The difficulty of attributing action to an individual scales with the number of people a system
must tell apart, and the published record tracks this scaling closely, even though almost no paper
frames it this way explicitly. At one actor, the problem largely disappears: golf-swing analysis
systems, single-athlete biomechanics tools, and gymnastics scoring aids operate on a single body
against a mostly static background, and because there is only one candidate for any detected
motion, identity is free. Boxing sits at a similar point on the axis when analysed one fighter at a
time, and hyperparameter-optimised convolutional models for boxing action recognition report
strong closed-set accuracy on exactly this kind of single-subject footage
\citep{kodathala2025sixsigma}. At two actors in direct opposition (tennis, padel, one-on-one
combat sports) a court's own geometry does much of the identity work for a system: a net fixes
which physical half of the frame belongs to which player, so the vision problem reduces to
detecting an event and reading off a coordinate, rather than distinguishing one body from another
among several. Consumer tennis-analysis products built on exactly this geometry publish accuracy
claims that team-sport products conspicuously do not, and the difference is best explained by the
geometry rather than by any difference in underlying model quality.

The moment a team, rather than a single opponent, is on one side of that same geometric split, the
guarantee weakens. Two teammates in a beach-volleyball pairing, two partners in a doubles match,
occupy the same half of the court and very often wear the same kit, so the net no longer settles
who is who within a side; a system must now actually distinguish two visually similar bodies, and
public accuracy claims for this exact setting are markedly harder to find than they are for
singles. Push the team size further, to the six players on a volleyball side or the five on a
basketball team, and identity is no longer localisable by geometry at all: any of several players
occupying overlapping regions of the frame could plausibly be the one who acted, uniforms are by
design identical within a team, and the market's own documentation (examined in detail in
Section~\ref{ssec:vendors}) shows every commercial system quietly reintroducing a human step at
exactly this point. Extend the frame once more, to an eleven-a-side football pitch or an American
football roster of comparable size, and a second axis of difficulty appears alongside crowding: the
playing surface is now large enough that a single fixed camera cannot keep every actor at a legible
resolution simultaneously, so the choice becomes between a wide shot in which nobody is readable
and a tight shot in which most of the field is out of view. The same tension between coverage and
legibility recurs, in a smaller and more tractable form, in the volleyball footage we study here,
where camera angle relative to the court, rather than field size, is the dominant constraint: a
single fixed camera's viewing angle measurably affects downstream tactical-recognition accuracy on
exactly this kind of footage, even without any change in field-of-view or resolution
\citep{cheng2023pathfinder}.

The academic sports-video literature mirrors this progression closely, but its benchmarks sit
almost entirely at the easy end of it. Multi-view, three-on-three basketball captured on a fixed
consumer camera is treated as a distinct, harder condition worth its own dataset precisely because
it departs from the broadcast norm \citep{matsune2025trackid3x3}. Tracking and re-identification
benchmarks built from broadcast basketball, football, and volleyball footage report strong headline
numbers for detecting and following people \citep{cui2023sportsmot,li2021multisports}, but those
numbers are earned on footage shot by professional camera operators, with multiple angles, replay
coverage, and rosters of players whose faces and gait a re-identification model can learn across
an entire season. Jersey-number recognition, the single most direct proxy for identity in team
sport, has been studied carefully enough that its own authors report the pixel-level limit
directly: on a corpus mixing broadcast hockey with a fixed, unzoomed university-gymnasium camera,
only a small single-digit percentage of player crops are legible enough to attempt a reading at
all, and the authors attribute this explicitly to the fixed camera's lack of pan and zoom rather
than to any deficiency in the reading model itself \citep{koshkina2024jersey}. A companion study
extending the same pipeline to long-term tracking finds that appearance-based re-identification
confidence decays from near-certainty to near chance within a few hundred frames of an actor
leaving and re-entering the frame, which is the empirical reason a system cannot simply
re-recognise a player by appearance once it has lost track of them \citep{koshkina2025longterm}.
Game-state reconstruction, which asks a system to place every athlete on a schematic map together
with their team, role, and jersey number from broadcast video, reports headline accuracy in the
low twenties out of a hundred for the complete task even with professional camera work
\citep{somers2024gsr}, and successive editions of the same challenge show the score improving by
under a single point year over year, with organisers naming jersey-number recognition explicitly
as the bottleneck component \citep{giancola2025soccernet}. The most recent instance of this
challenge line reframes the task specifically around attributing an action to the player who
performed it, rather than merely tracking bodies, and the winning entry still recovers well under
two-thirds of instances correctly under a protocol that supplies ground-truth tracks the system
does not have to discover on its own \citep{giancola2026soccernet}. A separate basketball
benchmark built explicitly to test whether a system can say who did what and when, again on
broadcast footage with an official play-by-play feed available to align against, likewise reports
that state-of-the-art multimodal systems remain well short of reliable event-to-player attribution
\citep{hu2026basketball}. Taken together, this literature already contains the finding our own
measurements confirm on much harder footage: detecting that something happened is close to solved;
saying who did it is not, and the gap between the two grows, rather than shrinks, as the number of
candidate actors increases and the camera work becomes less forgiving.

\subsection{What the market actually ships}
\label{ssec:vendors}

A reader could reasonably assume that a gap this well documented in the research literature has
simply not yet reached commercial products, and that some vendor, somewhere, has already closed
it. The public record of the sports-technology market does not support that assumption, and it is
worth setting out carefully, because every claim below is drawn from a vendor's own published
documentation rather than from a competitor's characterisation of it.

The clearest historical data point is a company that no longer exists. A widely used sports-film
platform built its early per-player statistics on a network of roughly two thousand paid human
reviewers, who tagged game film by hand for a fee of about fifteen to twenty dollars per game and
a turnaround measured in hours rather than seconds; its own founder, discussing the company's
approach publicly, stated plainly that ``humans are still better than computer vision'' for this
task \citep{si2015krossover}. The company was acquired specifically for that reviewer workforce
\citep{hudl2019krossover}, and the product built on it was discontinued a little over a year later. Nothing in that history is a criticism of
the people involved; if anything, it is evidence that the market tried the economics of scaling
human review to this problem and found the arithmetic did not close, which is itself informative
about how hard the underlying perception problem is.

The products operating today tell a more nuanced but ultimately consistent story. A major
film-analysis platform markets its automated offering with language promising an end to manual
tagging, and in the same breath, on the same support page, states that the core of the service
``relies on a team of trained analysts who manually tag every play, event, and statistic in a
video'' \citep{hudl2026assistfaq}. The same platform's automated tier for volleyball specifically
requires the customer to supply jersey colours, the list of athletes who played, and the final
score by hand before any processing begins, and its own documentation states that any player not
already listed on a pre-submitted roster will have their statistics attributed to an
``unknown athlete'' placeholder rather than to a name \citep{hudl2026assistvolleyball}. Another
volleyball-specific platform claims full automated per-player attribution, kills, digs, blocks,
aces, assists, reception, and attack efficiency, from an ordinary phone camera, with no published
per-touch accuracy figure for any of it \citep{sportsvisio2026volleyball}. Its own capture guidance
is, in this respect, the most directly informative document we found anywhere in the market: it
instructs a customer to film from a position that is ``stable, elevated, and behind the end line so
the full court and all six rotations stay in frame,'' to keep ``jersey numbers readable whenever
possible,'' and states plainly that ``poor capture habits cost more accuracy than cheap hardware
does'' \citep{sportsvisio2026capture}. Every element of that customer-facing checklist names exactly
the two variables Sections~\ref{sec:setting} and~\ref{sec:discussion} measure directly and show to
dominate identity accuracy in our own corpus, camera axis and digit legibility, which is
independent corroboration of this paper's central empirical claim from a vendor with every
commercial incentive to claim the opposite, that its system works regardless of how a customer
happens to point their phone. A phone-based volleyball analysis product markets itself as requiring
``no stat crew, no tagging,'' while its own frequently-asked-questions page states that players on
the same team must wear distinguishable numbers for the system to function and that submitting a
player's correct jersey number before the season begins is, in the vendor's own words, ``crucial''
to accurate attribution \citep{balltime2026faqs}. A soccer-analysis platform attributes touches to players not through
computer vision at all but through a wearable sensor worn on the player's own body, and states
outright, in its athlete-facing materials, that a player who forgets the sensor does not appear in
the statistics for that match \citep{trace2026howitworks}. Several installed-camera systems require a professional or
vendor-performed calibration step (in one documented case, up to two working days) after which
the camera rig is explicitly instructed not to be moved again \citep{spiideo2026calibration}, trading the flexibility of a
handheld or repositionable camera for a fixed, pre-verified geometry closer to the broadcast
condition the underlying models were most likely trained toward. One competitor's own public
comparison of automated sports-statistics vendors makes the point more bluntly than any academic
source could: it argues explicitly that the right question to ask a vendor in this category is not
whether its output is accurate, but whether a customer can see and correct what it got wrong
\citep{smashvision2026comparison}, which is itself a tacit admission that the accurate case cannot
be assumed.

None of this is offered as an accusation against any individual company; building a product that
customers can actually use, on a budget a youth sports club can actually pay, is a legitimate and
difficult engineering problem, and moving the labour of identity resolution onto a human (an
analyst, an installer, a parent filling in a roster, the athlete's own wearable sensor) is a
coherent way to ship something that works today. The pattern is nonetheless clear and worth
stating plainly, because it is the empirical premise this paper tests directly rather than assumes:
every commercially available system we could find documentation for that claims automatic
per-player attribution in a team sport requires, at some stage between the camera and the box
score, either a fixed and carefully positioned rig, a pre-registered roster, a wearable device, or
a person watching the footage. The one apparent exception in the literature (a provider offering
markerless per-player tracking with no customer-side hardware or labour at all) operates
exclusively on professional broadcast feeds of a small number of elite competitions, and an
independent validation of comparable broadcast-derived, computer-vision tracking software against
ground-truth optical tracking at a professional match reports position error ranging up to double
digits of metres depending on feed type, and finds that most of that error is attributable not to
imprecision when a player is detected but to the player not being detected at all
\citep{crang2025concurrentvalidity}. No public claim we could
verify describes a system that takes an ordinary, single, unattended camera and an unprepared
roster and returns reliable per-player statistics without a human closing the gap somewhere in the
pipeline. This paper measures, in detail, what closing that gap with computation alone actually
costs, and where it succeeds and fails.

Two vendors illustrate a related but distinct move: introducing a smarter camera rather than a
smarter reading of the video. One permanently mounted system ships sport-specific mounted angles,
a wired network connection, and dedicated tracking algorithms separately tuned for American
football, soccer, lacrosse, basketball, volleyball, and wrestling \citep{hudl2026focus}; a second,
phone-based system locks a specific iPhone model into a proprietary case and auto-pans and zooms
across soccer, futsal, basketball, and volleyball \citep{ballercam2026howitworks}. Both are genuine
engineering achievements at the capture problem, keeping the ball and players framed without a human
operator, and neither claims to solve the problem this paper measures: the mounted system's own
sport-specific pages describe camera angles and recording duration, not per-player accuracy, and the
phone-based system's advertised ``advanced stats'' derive from scoreboard detection and AI-flagged
plays rather than from a published per-player attribution figure. A better camera answers where to
point; it does not answer who touched the ball, and the two problems do not share a solution. We
build on the second side of that split deliberately: this paper's system is software that reasons
over video a program already has, from whatever camera it already owns, rather than a piece of
hardware a program must first buy, install, and in the mounted case wire into its network. Given the
budget reality established in Section~\ref{sec:intro}, a capital purchase and an installation are
themselves a barrier a majority of the programs this paper is motivated by cannot clear, independent
of whether the camera behind that purchase can frame a play well.

\subsection{Video understanding beyond sport}

The identity-attribution gap documented above for sport is a specific instance of a more general
observation in the video-understanding literature: benchmarks that reward describing a scene are
not the same benchmarks that would reward attributing an action within that scene to a specific
individual, and strong performance on the former predicts little about the latter.
Table~\ref{tab:benchmarks} makes this concrete with published numbers rather than assertion. Current
video-understanding benchmarks do not move together: a broad, long-context benchmark is answered at
a high absolute level, above four correct answers in five \citep{wu2024longvideobench}, while a
benchmark built specifically to stress fine-grained temporal reasoning, counting, direction, causal
ordering, shows the best reported model landing 57 points below a measured human baseline on the
same task \citep{shangguan2024tomato}, and a benchmark targeting fine-grained temporal event
ordering specifically lands at a comparably low absolute score, though its authors do not report a
human baseline to measure the gap against directly \citep{cai2024temporalbench}. A newer, stricter
scoring of general video question answering lands in between \citep{fu2026videommev2}. A fifth
benchmark makes the point this paper cares about most directly, rather than as a byproduct of
difficulty: its authors built it specifically because standard single-turn video QA had reached
roughly 90\% on a widely used leaderboard, a level its own authors judge no longer informative
about a model's real capability, and reformulated the task as multi-turn and tool-augmented,
requiring a model to look again, gather further evidence, and integrate it across turns rather than
answer once from a fixed context. Every frontier model tested, including systems otherwise reported
at the top of single-turn leaderboards, scores under 60\% once the task is posed this way
\citep{zhang2608videogaia}. That is, to our knowledge, the clearest existing evidence from outside
our own corpus that single-pass and agentic video understanding are different capabilities rather
than the same one at different levels of effort, exactly the split Section~\ref{sec:results}
reports between single-pass and agentic performance on our own footage.

These four benchmarks, LongVideoBench, TOMATO, TemporalBench, and the newer Video-MME-v2, are not
an arbitrary sample: they are among the most widely cited current tests of exactly the
general-purpose video-understanding capability this paper's introduction and abstract invoke, and
the models that top their public leaderboards are drawn from the same general-purpose model
families, Gemini and GPT, that we evaluate directly on our own footage in Table~\ref{tab:readers}
and throughout Section~\ref{sec:results}, not a different or weaker substitute chosen to make our
own results look worse by comparison. This is the comparison the rest of this section makes
concrete, not a hypothetical extrapolation from a public leaderboard to a sport nobody benchmarked,
but the same families of model, evaluated by us, on footage we control and ground truth we label,
at exactly the point their own public benchmark scores predict they should start to struggle.

\begin{figure*}[t]
\centering
\includegraphics[width=0.82\textwidth]{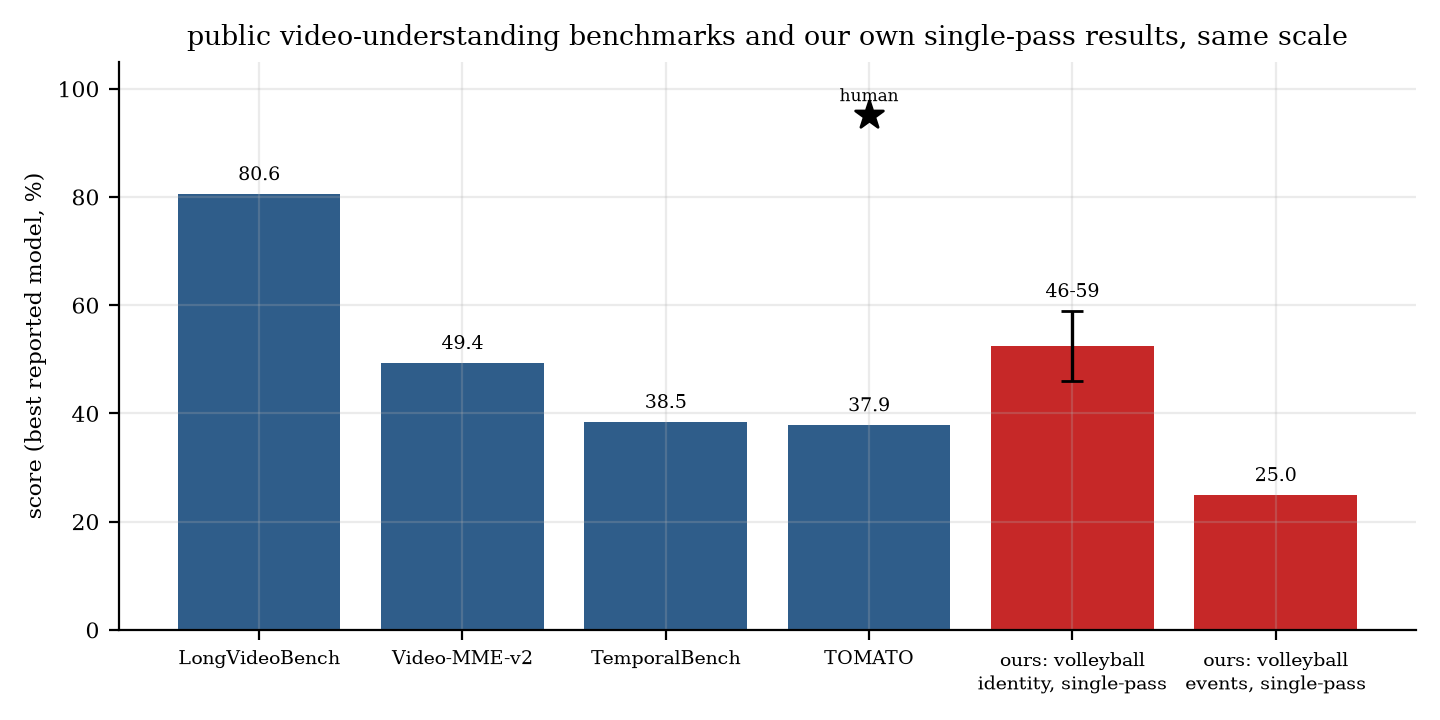}
\caption{Public video-understanding benchmark scores (blue, best reported model per
Table~\ref{tab:benchmarks}) against our own single-pass volleyball results (red), plotted on the
same 0--100 scale. Stars mark a published human baseline where the benchmark's own authors report
one. Our identity range comes from the nine single-pass readers in Table~\ref{tab:readers}; our
event-detection figure is the constrained, non-agentic configuration reported in
Section~\ref{ssec:wholematch}. Both fall inside the range the hardest, most fine-grained public
benchmarks already predict for this class of task, not below it.}
\label{fig:benchvsours}
\end{figure*}

A benchmark score reported in isolation does not say whether a task is solved, so these gaps need
a sense of scale. A ten-to-twenty-point gap to a measured human baseline is
roughly the range in which many mature computer-vision tasks still sit while nonetheless being
treated as usable, deployed capabilities; a fifty-point-or-larger gap, the range TOMATO and
TemporalBench report, is closer to the range in which a task is still treated as substantially
unsolved research rather than a deployment candidate. Our own single-pass results on volleyball sit
at this second, wider scale of gap, not the first: the nine commercial readers in
Table~\ref{tab:readers}, each given one bounded prompt over one clip with no agentic re-examination
or second look, span identity accuracy from 0.46 to 0.59 against a human-labelled ground truth a
coach would expect to be close to exact, and the same single-pass configuration finds only about a
quarter of a match's true events at all before any agentic reasoning is added
(Section~\ref{sec:results}). Both figures land inside the range the benchmarks in
Table~\ref{tab:benchmarks} already predict for this class of task, which is the point of citing them
here: our results are not an anomaly specific to amateur volleyball, but close to what the field's
own general-purpose benchmarks already say to expect once a task stops being coarse scene
description and starts requiring exact, instance-level temporal attribution.
Figure~\ref{fig:benchvsours} puts both sets of numbers on one axis rather than leaving the reader
to hold four benchmark scores and two of our own in their head at once.

The pattern
across these four benchmarks, general and coarse scores high, fine-grained and temporal scores low,
is exactly the split this paper's own stage-by-stage results (Section~\ref{sec:results}) reproduce
directly, rather than merely echo: the same reasoning-capable models we evaluate as whole-match
reasoners and as jersey readers in Section~\ref{sec:results} are, respectively, strong at the
coarse, scene-level judgement segmentation requires and weak at the fine-grained, small-object
judgement identity requires, on our own footage and our own held-out labels rather than on a
published leaderboard. Self-supervised
video world models trained at very large scale report strong transfer to action classification and
short-horizon prediction \citep{bardes2024vjepa,assran2025vjepa2}, yet the same model family's own
authors describe its feature maps, on closer inspection, as retaining only fragmented local spatial
structure rather than the precise, instance-level detail that fine-grained attribution would
require \citep{murlabadia2026vjepa21}. A direct probing comparison of two prominent self-supervised
video and image backbones on sports action data (work that one of us conducted prior to, and in
service of, the present study) finds that neither representation, used off the shelf, carries
enough spatially precise information to support the kind of fine-grained per-actor discrimination
this paper requires \citep{kodathala2025temporal}, a finding this paper's own frozen-feature
probing experiment (Section~\ref{sec:results}) replicates on volleyball footage specifically.
Benchmarks built to stress small-object and fine-grained visual perception in frontier
vision-language models report accuracy in the low teens even for the strongest available systems,
and find, counter-intuitively, that giving a model more of the video (more frames, a longer
temporal context) can \emph{reduce} its accuracy at exactly this kind of task rather than improve
it \citep{li2604videozerobench}. Sports-specific reasoning benchmarks report the same pattern under
a different name: models can often answer a question about foul type or game state correctly while
their grounding of that answer to a specific location in the frame remains close to unusable
\citep{xia2511sportr,xu2026refereebench}. The broader agentic-vision literature on visual search and
crop-and-zoom reinforcement learning arrives at a closely related conclusion from a different
direction: the accuracy gained by teaching a model to look more carefully at a region of interest
is dominated by what the model already knew before it looked, rather than by information the
zooming action itself recovers \citep{ma2026visiontooluse}, and an iterative visual self-correction
loop shows a closely related failure mode, in which any apparent benefit disappears once a
deployable stopping rule replaces the benefit of hindsight \citep{tripathy2026selfcorrectionmirage}.
Multi-step and multi-agent language-model pipelines show a parallel, non-visual version of the same
phenomenon, with detection accuracy for an injected error falling substantially across successive
pipeline stages even when each individual stage is independently capable
\citep{singh2026hallucinationsnowball}, and a controlled comparison finds that a far simpler
two-call pipeline can outperform an elaborate five-agent one, precisely because chaining
independent decisions compounds their errors rather than correcting them
\citep{prajapati2026twocalls}. Our own findings in Section~\ref{sec:failure} sit inside this
literature rather than apart from it: the reason identity resists every method we test is not that
our footage is unusually difficult in some idiosyncratic way, but that it sits at the specific
combination of small-object, fine-grained, multi-actor conditions that this broader body of work
has already shown, across several unrelated domains, to be exactly where video understanding
currently stops.

\begin{table}[t]
\centering
\caption{Reported scores on five current video-understanding benchmarks, as published by each
benchmark's own authors. Scores are not comparable across rows: each benchmark uses its own task
mix, scoring rule, and difficulty profile. What the table shows is not a ranking but a spread ---
general, coarse-grained benchmarks are answered at a level approaching or exceeding informal human
performance, while benchmarks built specifically to stress fine-grained temporal reasoning, or to
require agentic, multi-turn use of a video rather than a single pass over it, show a wide,
persistent gap.}
\label{tab:benchmarks}
\begin{tabular}{lp{3.6cm}rr}
\toprule
Benchmark & What it stresses & Best reported model & Human baseline \\
\midrule
LongVideoBench & long-context retrieval over extended video & 80.6\% & not reported \\
Video-MME-v2 (2026) & broad video QA, stricter group scoring & 49.4\% (group score) & not reported \\
TemporalBench & fine-grained temporal event ordering & 38.5\% & not reported \\
TOMATO & visual temporal reasoning (counting, direction, causality) & 37.9\% & 95.2\% \\
VideoGAIA (2026) & agentic, tool-augmented, multi-turn video use & under 60\% & not reported \\
\bottomrule
\end{tabular}
\end{table}

\section{Corpus and Setting}
\label{sec:setting}

Everything measured in this paper was measured on the same corpus: 66 amateur volleyball matches,
played by youth and school teams and filmed on a single fixed camera positioned by a coach, a
parent, or a school's own gymnasium rig rather than by a trained camera operator. Table~\ref{tab:corpus}
summarises its scale. Every one of the 46{,}648 contacts in the corpus was labelled by a person
watching the film: which team touched the ball, what the contact was, and, for the majority of
contacts, which numbered player made it. This is, to our knowledge, the largest per-touch labelled
volleyball corpus assembled outside a professional data-collection operation, and its scale is what
makes the negative results in this paper credible rather than anecdotal: a method that fails on
five matches might simply have had a bad afternoon, but a method that fails consistently across
sixty-six, spanning multiple gymnasiums, camera positions, and years of footage, is failing for a
structural reason.

\begin{figure*}[t]
\centering
\includegraphics[width=0.92\textwidth]{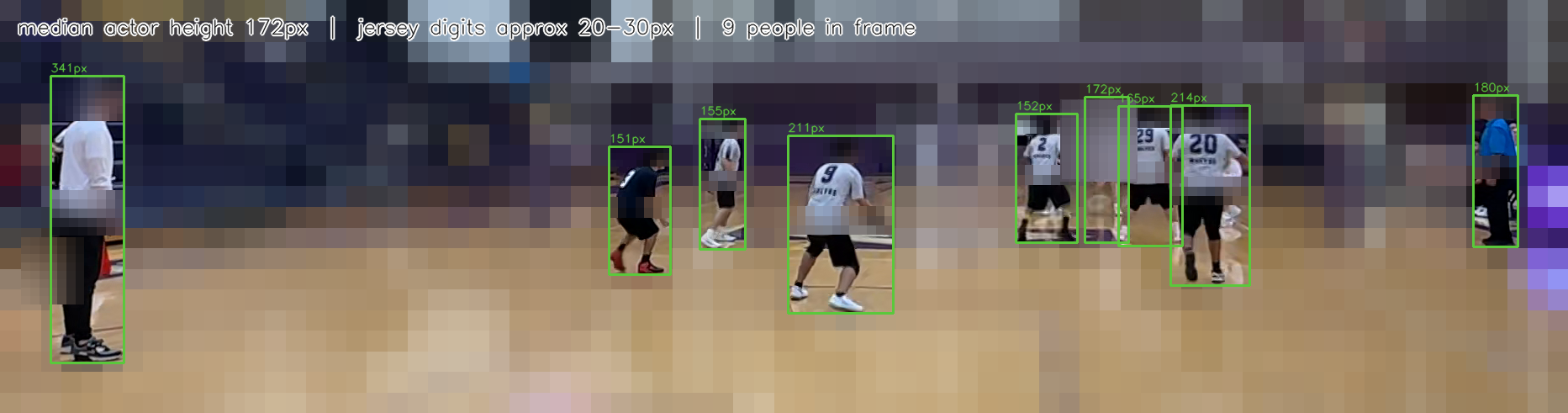}
\caption{A single representative frame from the corpus, with every tracked player boxed and its
pixel height labelled. Faces, bystanders, and venue-identifying signage have been removed as
described in Section~\ref{sec:statements}; the boxes and measurements are otherwise exactly as
extracted from the source footage. Nine players are visible simultaneously, close to the corpus
median for a live rally, and the median player height in this frame is representative of the
corpus as a whole.}
\label{fig:context}
\end{figure*}

\begin{figure}[t]
\centering
\includegraphics[width=\columnwidth]{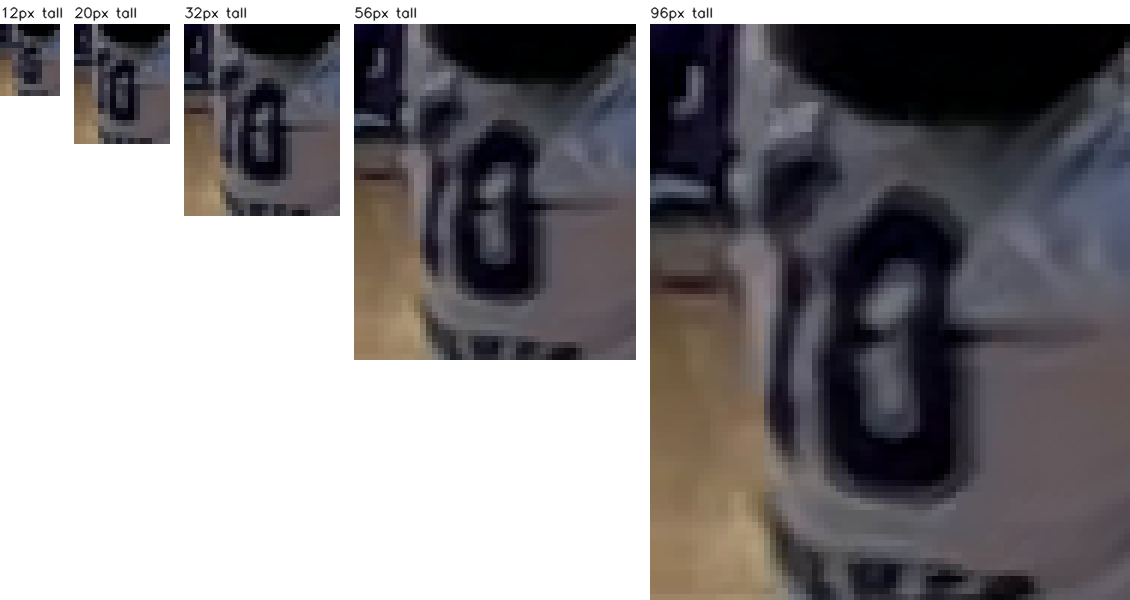}
\caption{A single jersey-number region from the corpus, downsampled to five representative pixel
heights and upsampled back for comparison. Below roughly twenty pixels the digit is no longer
reliably distinguishable from adjacent numerals, independent of any model's capability; this is
an optical limit rather than a modelling one, and it is the direct evidence behind the resolution
findings reported in Sections~\ref{sec:results} and~\ref{sec:failure}.}
\label{fig:opticallimit}
\end{figure}

We deliberately did not curate this footage toward conditions that would make automated analysis
easier. Figure~\ref{fig:context} shows a single representative frame with every tracked player
boxed and their pixel height labelled; identifying details of the venue, the bystanders, and any
face have been removed for privacy reasons described in Section~\ref{sec:statements}, but the boxes
and measurements are exactly as extracted from the original footage. The median player height
across the corpus is under two hundred pixels, and a jersey digit occupies a small fraction of that
height, typically between fifteen and thirty pixels tall, well below the range that optical
character recognition guidance for industrial and document imaging considers reliable. This is not
an unusual or worst-case frame; it is a typical one, chosen because it happens to contain nine
players simultaneously, which is close to the corpus median for a live rally. Figure~\ref{fig:opticallimit}
makes the consequence of that pixel height concrete: the same jersey digit, downsampled to a range
of heights spanning the corpus distribution, stops being reliably distinguishable from a
neighbouring digit well before it reaches the lower end of that range, independent of any reading
model's capability.

\begin{figure}[t]
\centering
\includegraphics[width=\columnwidth]{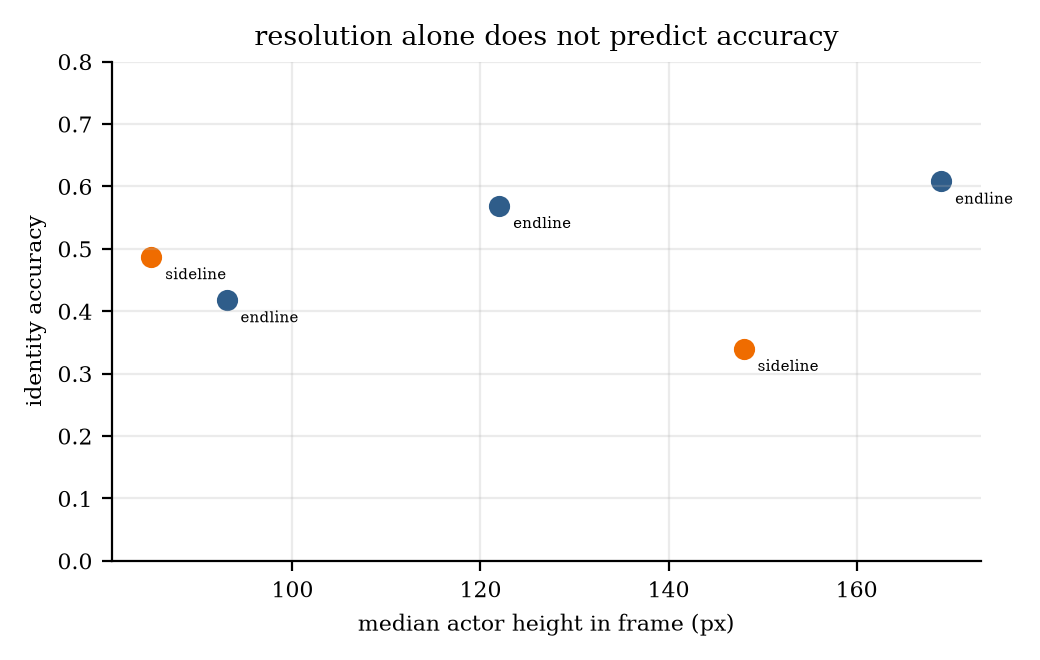}
\caption{Identity accuracy against median actor height in frame, split by camera axis (endline
versus sideline). Accuracy does not order monotonically with height alone; the combination of
height and viewing angle is the better predictor, discussed in Section~\ref{sec:setting}.}
\label{fig:heightaxis}
\end{figure}

\begin{figure*}[t]
\centering
\includegraphics[width=0.78\textwidth]{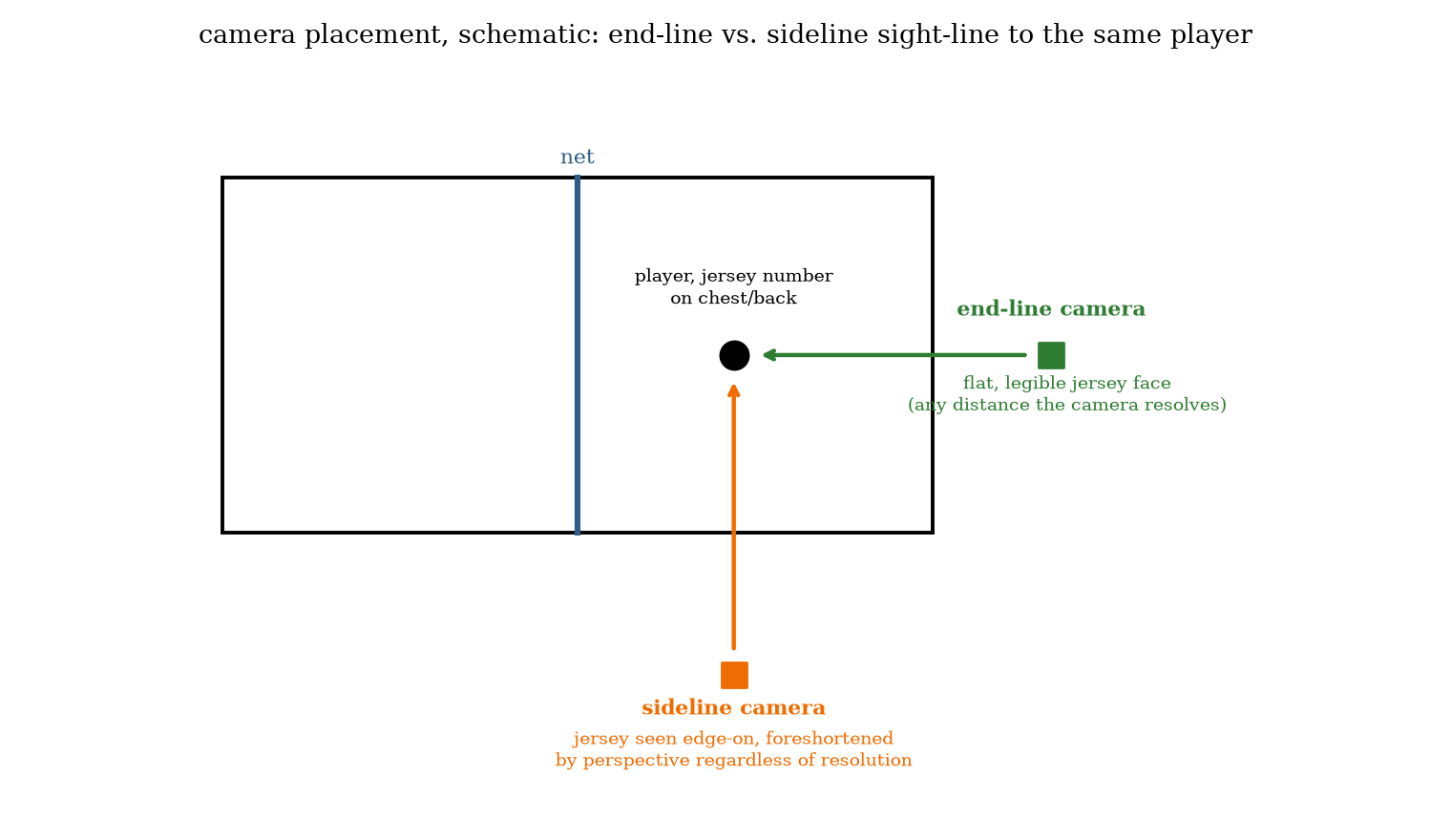}
\caption{The geometry behind Figure~\ref{fig:heightaxis}, drawn rather than only described. An
end-line camera looks down the length of the court, so a jersey number stays close to
fronto-parallel to the lens no matter how far the player is from the net; a sideline camera looks
across the court's width, so the same jersey number is seen increasingly edge-on toward the far
side of the court, an effect no amount of resolution corrects.}
\label{fig:cameraplacement}
\end{figure*}

\begin{figure*}[t]
\centering
\includegraphics[width=0.98\textwidth]{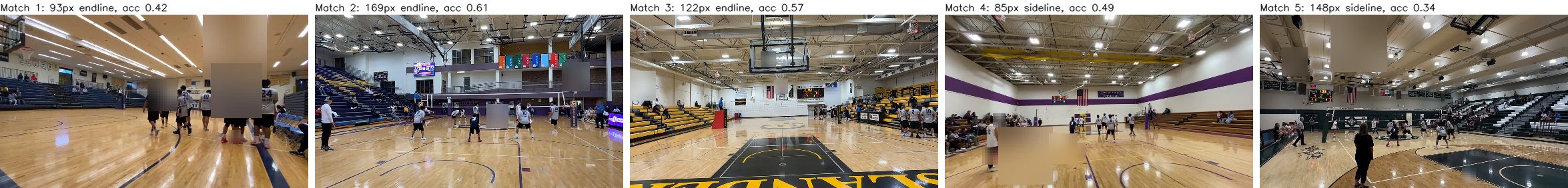}
\caption{Five real, unedited camera conditions from five different held-out matches, not one match
shown five ways: the same heterogeneity Figure~\ref{fig:heightaxis} plots numerically, seen
directly. Faces are blurred for privacy; framing, distance, and axis are exactly as filmed. Labels
give each match's median player height, camera axis, and measured identity accuracy
(Table~\ref{tab:pergame}).}
\label{fig:camerasamples}
\end{figure*}

The corpus is also not homogeneous, and that heterogeneity turns out to matter more than any single
property in isolation. Of the matches in our held-out evaluation set, five carry per-player jersey
ground truth and anchor the per-match identity results reported in Section~\ref{sec:results} and
Table~\ref{tab:pergame}. Those five span a median player height from under ninety to over a
hundred and seventy pixels and include both cameras placed along the sideline of the court, which
shows players in profile, and cameras placed behind the end line, which shows the backs and fronts
of the two teams respectively; Figure~\ref{fig:camerasamples} shows what these five conditions
actually look like side by side, rather than only as summary statistics. Figure~\ref{fig:heightaxis} plots identity accuracy against both quantities
jointly, and the picture that emerges is not the simple monotone relationship a reader might
expect: the match with the largest players in frame is not the one with the highest identity
accuracy, and a smaller-player match filmed from behind the end line outperforms a larger-player
match filmed from the sideline. Figure~\ref{fig:cameraplacement} shows the geometry behind that
reason directly. The reason is that an end-line camera shows the flat, legible face
of a jersey number, while a sideline camera shows it edge-on, foreshortened by the angle regardless
of how many pixels the player otherwise occupies. The clearest single illustration of this is a
direct crossover in our own results: the larger of our two sideline matches, with a median player
height of 148 pixels, scores the lowest identity accuracy of any match we measured, 0.340, while an
end-line match with barely more than half as many pixels per player, 93, scores higher at 0.418,
and the two end-line matches with still more pixels, 122 and 169, score 0.569 and 0.608
respectively (Table~\ref{tab:pergame}). More pixels, in this specific and reproducible sense,
bought the larger sideline match nothing: the camera's axis relative to the court, not its distance
from it, is the variable actually load-bearing for this stage. This is also why we treat camera placement as a
design variable a practitioner can act on rather than a fixed property of a venue to be modelled
around: unlike sensor resolution, which a customer typically cannot change without buying different
hardware, where to stand with the camera a customer already owns costs nothing, and, on this
evidence, matters more than what that camera is. Deployability, in other words, is at minimum a
two-dimensional property, resolution and viewing angle together, and any claim that a system
works reliably ``at 1080p'' or ``above some pixel threshold'' is, on this evidence, an incomplete
description of the conditions that actually determine whether it works.

Two further properties of the corpus are worth stating explicitly because they recur throughout the
paper's analysis. First, the game clock the corpus was labelled against runs to one-second
resolution, which is coarser than the sub-second precision a video model can, in principle, resolve
a contact to; we account for this explicitly in our evaluation protocol (Section~\ref{sec:results})
by matching predicted and true events within a tolerance window rather than requiring exact
timestamp agreement, and separately confirm that the great majority of contacts have an acoustic
onset, the sound of the ball being struck, within a fraction of a second of their labelled time,
which rules out systematic labelling drift as an explanation for the identity results that follow.
Second, every held-out result in this paper is reported on a frozen split of matches that were
never used to develop or tune any component of the system, a discipline that is unfortunately not
universal in the sports-video literature and that we discuss further, together with our full
evaluation protocol, in Section~\ref{sec:results}.

\begin{table}[t]
\centering
\caption{Corpus and the two invariants that anchor the rule-based stages of the pipeline. Both
are measured over the full held-out corpus, not assumed.}
\label{tab:corpus}
\begin{tabular}{lr}
\toprule
Quantity & Value \\
\midrule
Matches & 66 \\
Labelled contacts & 46,648 \\
Rallies & 8,251 \\
Contacts carrying a jersey number in the label & 26,345 (56.5\%) \\
\midrule
Every rally begins with exactly one serve & 0.9998 (8,249/8,251) \\
The next rally is served by the team that won the last & 0.9996 (8,048/8,051) \\
\bottomrule
\end{tabular}
\end{table}

\section{From Video to a Box Score: Seven Stages}
\label{sec:stages}

A finished box score is the end of a chain of decisions, not the output of a single inference.
We decompose that chain into seven stages, each with its own inputs, its own notion of a correct
answer, and, as Section~\ref{sec:results} shows, its own winning approach. We describe the stages
here in the order information flows through them; Section~\ref{sec:paradigms} then describes the
four ways we implemented and compared solutions to each.

\textbf{Segmentation} divides a continuous video into the individual plays (rallies, in
volleyball's terminology) that make up a match, discarding the dead time between them. A play
boundary is the coarsest unit any downstream stage depends on: an error here does not merely
mislabel one event, it can attribute an entire sequence of contacts to the wrong play or lose them
to the gap between two mis-drawn boundaries.

\textbf{Event spotting} finds the individual contacts within a segmented play: the moments at which
a player touches the ball. This is a temporal localisation problem at a much finer grain than
segmentation, typically requiring the system to place a contact within roughly a second of its true
time to be useful downstream.

\begin{figure*}[t]
\centering
\includegraphics[width=0.98\textwidth]{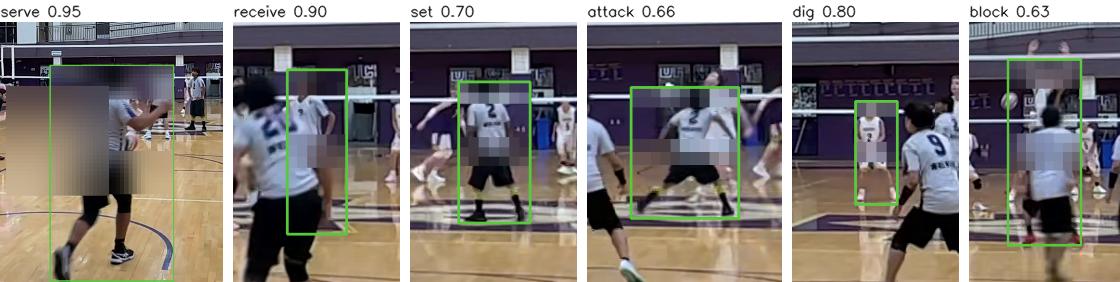}
\caption{Six real contacts from our corpus, one per action label, each shown with the tracked
candidate our own event spotter and reader selected (green box) and that model's own confidence
score for the label shown. These are not staged or synthetic examples; each is a genuine timestamp
and track drawn directly from the pipeline's own output on held-out footage, illustrating what
Section~\ref{sec:stages}'s action vocabulary looks like in practice rather than only in
definition.}
\label{fig:actionexamples}
\end{figure*}

\textbf{Action classification} labels what kind of contact each spotted event was (a serve, a
dig, a set, an attack, a block) using motion and context around the contact rather than the
contact instant in isolation, since the instant itself is often the most visually ambiguous moment
of the action. Figure~\ref{fig:actionexamples} shows six genuine, unstaged examples, one per
label, each with the tracked candidate and confidence score our own pipeline actually produced.

The action vocabulary we classify into is fixed by the sport rather than chosen by us: a
\emph{serve} puts the ball in play; a \emph{reception} is the first contact on an opponent's
serve; a \emph{set} is a contact, typically the second in a team's sequence, intended to position
the ball for an attack; an \emph{attack} is a contact intended to send the ball into the
opponent's court to end the rally; a \emph{block} is a contact made at the net against an
opponent's attack; and a \emph{dig} is a defensive contact that keeps an opponent's attacked ball
in play, a category that, by definition, excludes a serve receive and a ball played off a block
\citep{ncaa2026volleyballstats}. Two further terms sit one layer above this vocabulary and matter
specifically for how Section~\ref{sec:results} reports accuracy. A \emph{kill} is not a distinct
contact type; it is a judgement about how the rally containing an attack ended, an attack that was
not returned and won the rally outright, layered on top of the attack label itself. An
\emph{assist} is conventionally credited to whichever set immediately preceded a kill
\citep{ncaa2026volleyballstats}, which makes it a judgement about two already-attributed contacts
belonging to the same short sequence, layered on top of two separate attributions rather than one.
We do not report assist as its own statistic in Table~\ref{tab:statistics} for exactly this
reason: its accuracy is a product of the set's attribution accuracy and the kill's, both of which
Sections~\ref{sec:results} and~\ref{sec:limitations} already report as weak, and compounding two
weak attributions into a third, derived number seemed less honest than a row that could only ever
inherit, and worsen, the two weaknesses beneath it.

\textbf{Outcome determination} decides how a play ended: which team won the point, and, where the
sport's own definitions require it, what kind of terminal event ended it. This stage is unusual
among the seven in that its correct answer is frequently not directly observable from the video at
all, for reasons Section~\ref{sec:results} quantifies precisely.

\textbf{Team assignment} determines which of the two teams a given contact belongs to. This sounds
trivial and is not: a system that has correctly detected an event and correctly classified its
action can still credit it to the wrong side of the court, an error that (as we show) silently
cancels out in aggregate totals while devastating every per-team and per-player statistic
downstream of it.

\textbf{Player attribution} names the specific player who made a given contact. This is the
stage every other stage in the chain exists to serve, and it is, by a wide margin, the hardest of
the seven, for reasons that occupy the majority of Sections~\ref{sec:results} and~\ref{sec:failure}.

\textbf{Aggregation} sums the sequence of individually attributed events into the totals a coach,
player, or parent actually reads: total kills, service errors, digs, and the rest, at both the team
and the individual level. We treat this as a genuine stage worth measuring in its own right, rather
than as trivial arithmetic that follows automatically once the preceding six stages are correct,
because Section~\ref{sec:results} shows that arithmetic aggregation is a real and measurable source
of error even given perfect input.

Two properties of this decomposition matter for how the rest of the paper should be read. First, an
error at an earlier stage propagates to every later one, but a strength at an earlier stage does
not guarantee a strength at a later one: a system that segments matches flawlessly can still fail
completely at naming the player who acted within a correctly segmented rally. Second, and more
consequentially for the paper's central argument, each stage differs in how much of its correct
answer can be recovered from the fixed rules of the sport being played, independent of anything the
camera observed. A rally's outcome, for instance, is in volleyball almost entirely determined by
the identity of the team that serves the next rally, which the rules fix rather than leave open to
interpretation; a player's identity is not determined by any comparable rule at all. This difference
in how much of each stage's answer is \emph{given} by the domain, rather than requiring perception
to discover, is the single best predictor of which stages a given approach can solve well, and
Section~\ref{sec:results} organises its reporting around exactly this axis.

\section{Four Approaches}
\label{sec:paradigms}

We evaluate four qualitatively different ways of producing an answer at each of the seven stages
described in Section~\ref{sec:stages}. We describe each at the level of what kind of system it is
and what it was asked to do, rather than at the level of the specific prompts, thresholds, or
architectural configuration used to obtain our best results for it, for reasons stated explicitly
in Section~\ref{sec:statements}: this paper reports where each approach succeeds and fails, not the
calibrated recipe of the system we operate commercially.

\textbf{Video understanding and agentic reasoning.} A frontier, general-purpose vision-language model is given video or image
context drawn from the match and asked, in natural language, to produce the answer for a given
stage directly, to say where a rally begins and ends, to describe what a player did, to name a
jersey number. We evaluate this paradigm both in a single-call setting, where the model answers
from one prompt over a bounded span of footage, and in an agentic setting, where the model is
permitted to reason over multiple calls, request additional context, or reconsider an earlier
answer before committing to a final one. We treat commercially available reasoning-capable models
released across the period of this study as members of this paradigm, and report cost as the
metered spend a customer would actually incur running each one, at the provider's published rate,
on an identical workload.

\textbf{Computer vision and trained specialists.} Rather than asking one large model to reason over
raw video, this paradigm decomposes each stage into a purpose-built component: a detector that
finds people in a frame, a tracker that follows a detected person across frames, small models
trained specifically on this task's own labelled data to spot events, classify actions, or read a
jersey number from a sequence of crops, and a symbolic layer that encodes the sport's own rules as
explicit constraints on which sequences of events are legal. The trained components share a common
architectural shape rather than each being bespoke: a compact temporal model integrates frozen,
pretrained appearance features together with an explicit motion signal over a short window of
frames, rather than reasoning over a single instant, which is what lets it exploit exactly the
biomechanical redundancy described in Section~\ref{ssec:musclememory}; for the stages that require
distinguishing between several candidate people in the same frame, this temporal model is applied
along a region of interest pooled from each individual tracked candidate separately, rather than
over the frame as a whole, which is the architectural choice underlying the track-level identity
argument in Section~\ref{sec:failure}. This shape is a deliberate alternative to both of the
paradigms above: unlike video understanding through prompting, it never passes raw video through a language-modelling pathway at
all, so it is not bound by the fixed, coarse temporal sampling rate a general-purpose video API
imposes (Section~\ref{sec:results}); unlike the frozen video world-model paradigm below, its
features are fit directly to this task's own labelled data rather than left general-purpose, which
is what lets it resolve detail at the fine spatial and temporal scale this task needs rather than
the coarser scale a backbone trained for broad video understanding retains. We do not disclose the
specific backbone, window length, sampling rate, or layer configuration used to obtain our
production results, for the reasons given in Section~\ref{sec:statements}; every trained component
in this paradigm is nonetheless small by contemporary standards, on the order of a few million
parameters for the most complex of them, and every one is trained and evaluated on the frozen split
described in Section~\ref{sec:setting}, never seeing held-out matches during development.

\textbf{Video world models.} A third paradigm asks whether general-purpose, self-supervised video
representations, models trained at very large scale on unlabelled video to predict masked or
future content rather than trained on any labelled sports task at all, already contain enough
information to solve stages such as player attribution once a small classifier is fit on top of
their frozen features. We evaluate this by extracting features from a prominent self-supervised
video model and a prominent self-supervised image model over the same clips used to train the
task-specific specialists above, freezing the backbone entirely, and training only a lightweight
probe on top, following the same experimental design one of us used in an earlier, narrower
comparison of exactly these two representations on sports action data
\citep{kodathala2025temporal}. This paradigm answers a specific question the other three cannot:
whether the field's current best general video representations already carry the fine-grained,
instance-level information this task requires, independent of any task-specific training at all.

\textbf{Human annotation.} The fourth approach is the one the commercial market in this space
actually deploys, as documented in Section~\ref{ssec:vendors}: a person watches the footage and
writes down what happened. We include this not as an assumed ground truth against which to measure
the other three, but as a fourth method with its own measurable cost and latency, because treating
manual annotation as free, instant, or error-free obscures a comparison the market itself is
implicitly making every time it charges a customer more for a human-reviewed result than for an
automated one. Section~\ref{ssec:vendors} reports the clearest independently verifiable evidence we
found of what this actually costs in practice, drawn from a company that built its early product on
exactly this labour model; we did not independently measure the accuracy of manual sports
annotation against our own corpus and do not claim to.

For every stage and every paradigm where it was possible to do so without conflating cost with
capability, we held the underlying footage, tracks, and candidate crops constant across paradigms,
so that a reported difference in accuracy reflects a difference in the method's ability to use the
same evidence, not a difference in how much evidence each method was given. Where a paradigm
structurally requires different inputs, an agentic prompting system reasoning over raw video where
a trained specialist reasons over a cropped, tracked sequence, we report this explicitly rather
than treating the comparison as controlled.

% ---------------------------------------------------------------------------
% All figure floats. Tables are generated directly as complete \begin{table}
% environments by scripts/make_tables.py and \input separately.
% ---------------------------------------------------------------------------

\begin{figure*}[t]
\centering
\includegraphics[width=0.92\textwidth]{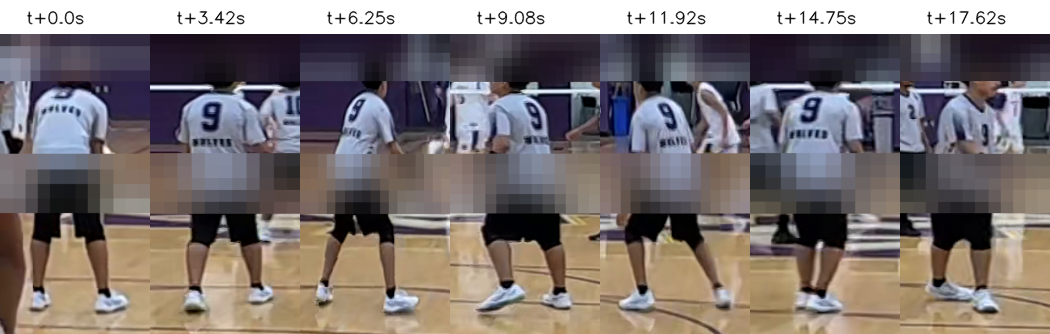}
\caption{The same tracked player, sampled at seven points across a single continuous play. The
jersey number is clearly legible at three of the seven sampled moments and unreadable at the
other four, where the player is turned, in motion, or partially occluded. This is the direct
visual evidence for the design argument in Section~\ref{sec:failure}: identity should be read
once, wherever across a track it happens to be legible, and propagated, rather than read only at
the instant of each individual contact.}
\label{fig:tracklet}
\end{figure*}

\begin{figure*}[t]
\centering
\includegraphics[width=0.98\textwidth]{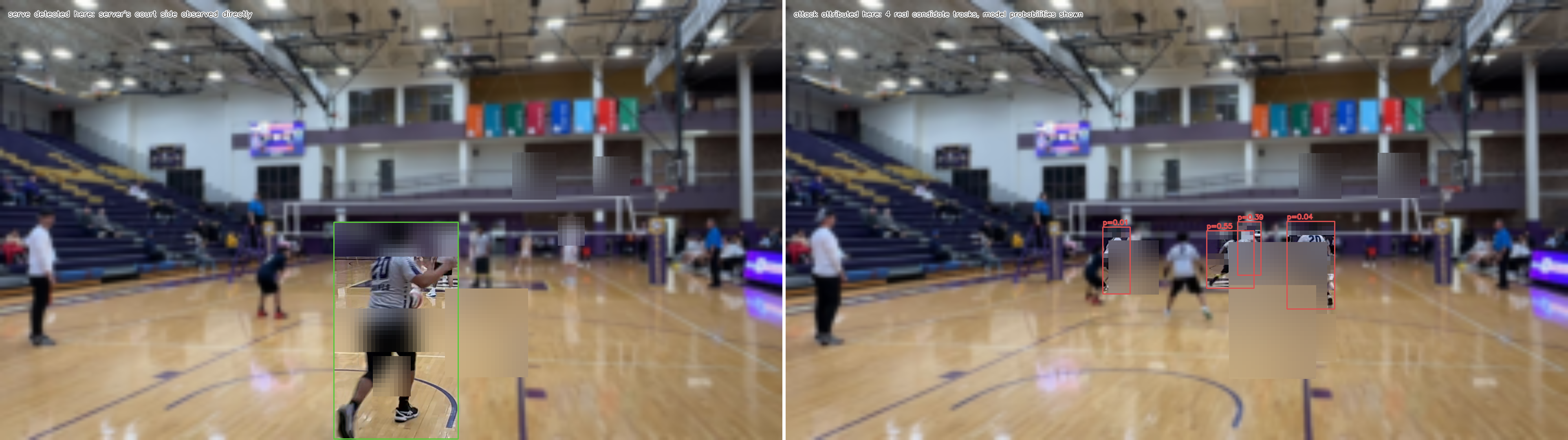}
\caption{Two real stages of the pipeline, shown rather than only described. \emph{Left:} team
assignment at a genuine serve: the server's physical position on the court is observed directly
(the mechanism behind the near-versus-far result in Section~\ref{sec:results}), not inferred from
which jersey colour is assumed to sit on which side. \emph{Right:} player attribution at a genuine
contested attack, with every real candidate track our own selection step considered and its actual
assigned probability labelled; the model split its confidence across two spatially close
candidates (0.55 and 0.39) rather than committing clearly to one, which is the concrete,
non-hypothetical shape of the selection failure decomposed in Figure~\ref{fig:decomposition}.}
\label{fig:teamattribution}
\end{figure*}

\begin{figure}[t]
\centering
\includegraphics[width=\columnwidth]{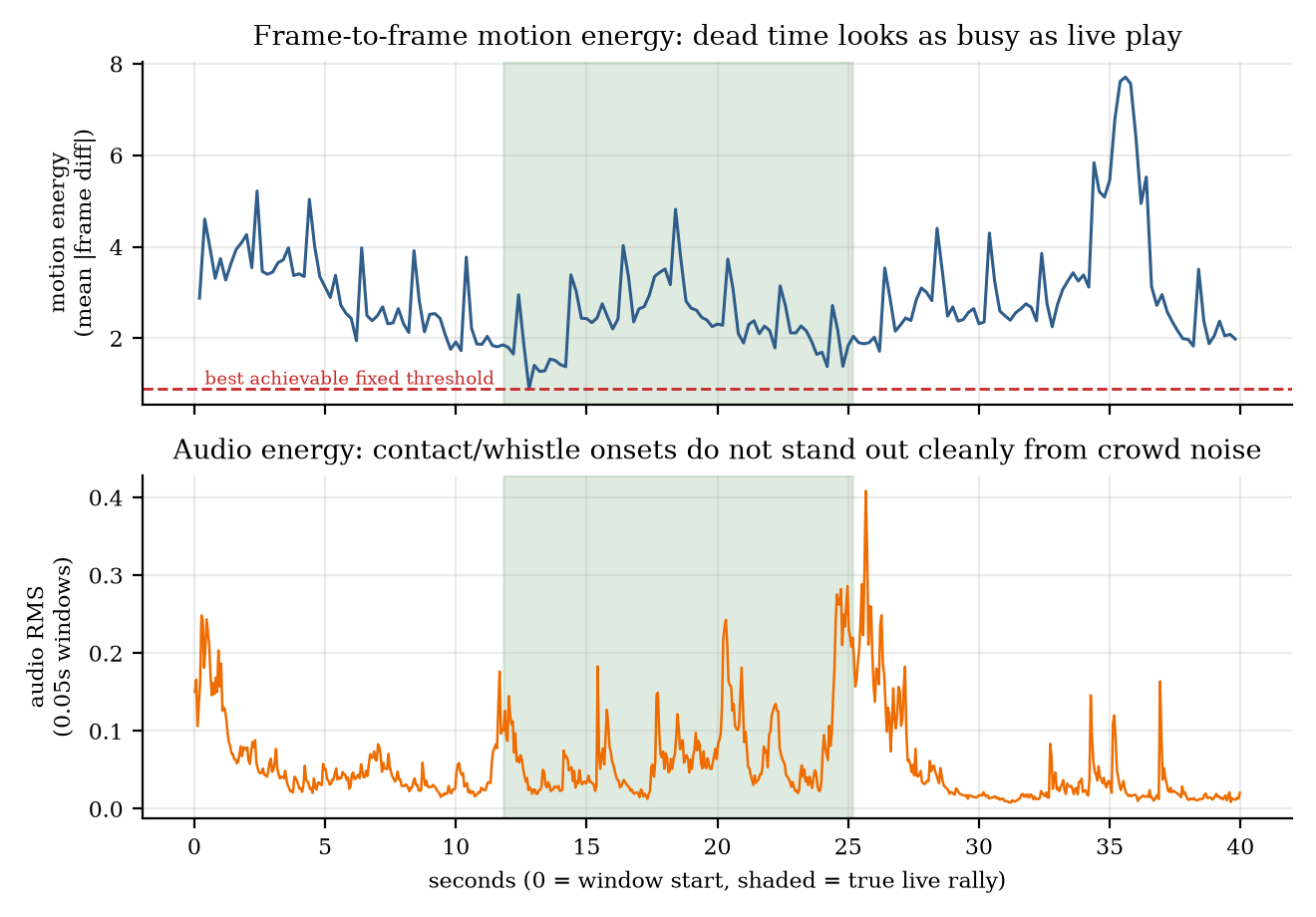}
\caption{Two classical, deterministic boundary signals computed directly from our own footage over
a real 40-second window (rally 20, true boundaries 695.17--708.5s, shaded), not a synthetic or
illustrative example. \emph{Top:} frame-to-frame motion energy (mean absolute pixel difference at
5 samples/second). The dashed line is the single fixed threshold that minimises total
misclassification error across this window; it still misclassifies roughly half of live-versus-dead
time, because dead time here (players resetting, walking to the bench) is, at points, more
motion-heavy than the live rally itself. \emph{Bottom:} audio RMS energy in 50ms windows over the
same span. Contact and whistle onsets sometimes produce a clear peak (near the rally's true end)
but comparable peaks recur throughout dead time and quiet stretches occur mid-rally, which is the
concrete picture behind the ``partial and unreliable'' audio result reported in
Section~\ref{ssec:segmentation}.}
\label{fig:motionaudio}
\end{figure}

\begin{figure*}[t]
\centering
\includegraphics[width=0.98\textwidth]{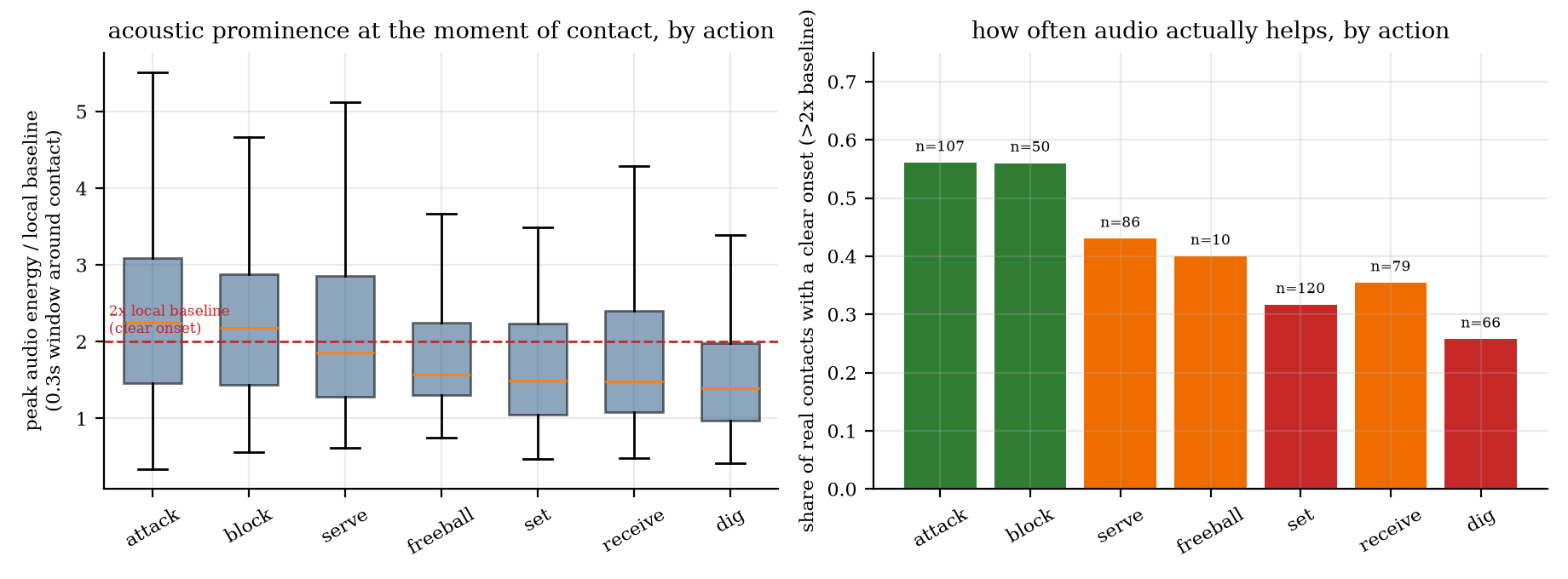}
\caption{Why audio helps at some contacts and not others, measured across every one of the
518 labelled events in this match rather than the single window in Figure~\ref{fig:motionaudio}.
For each real contact we compare the peak audio energy in a 0.3-second window around it to the
local baseline energy in the surrounding few seconds. \emph{Left:} the resulting distribution of
that ratio, by action; a ratio above 2$\times$ (dashed line) is a clear, usable onset. \emph{Right:}
the share of real contacts at that action type that actually clear that bar. Attacks and blocks,
the hardest-driven contacts, produce a clear acoustic signature a little over half the time; digs,
the softest and most defensive touch in the sport, clear it barely a quarter of the time. This is
the mechanistic reason a single audio-onset detector cannot be tuned to one threshold and used
across the action vocabulary: the events audio is worst at are exactly the events (digs, receives,
sets) that make up most of a match's contact volume (Table~\ref{tab:corpus}).}
\label{fig:audiobyaction}
\end{figure*}

\begin{figure*}[t]
\centering
\includegraphics[width=0.92\textwidth]{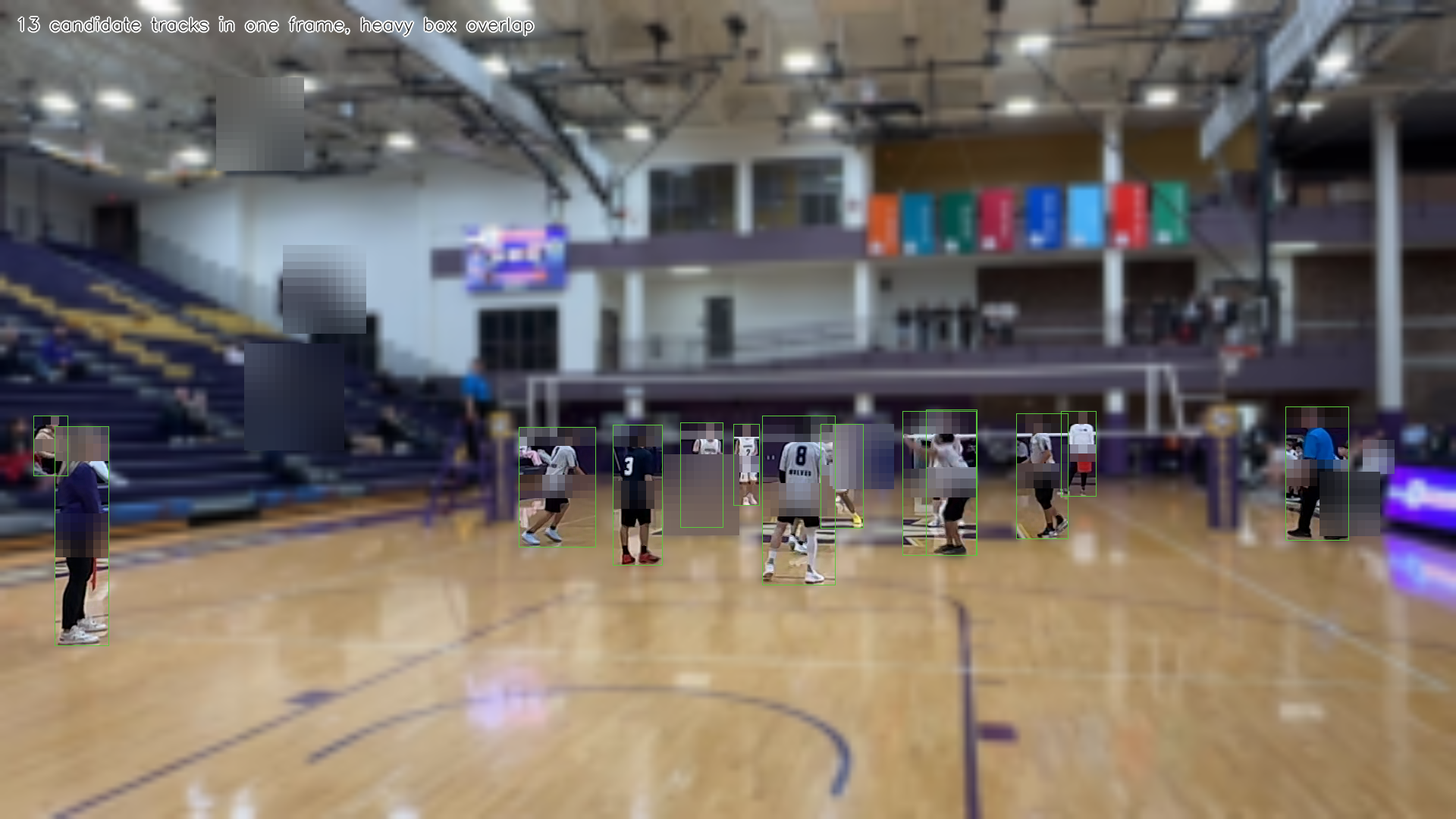}
\caption{The single most crowded moment we could locate in our own tracked corpus, found
programmatically by pairwise box overlap rather than chosen by eye: thirteen candidate tracks
occupy one frame, several pairs overlapping by more than two thirds of their area. This is the
concrete picture behind the identity-loss decomposition in Figure~\ref{fig:decomposition} and the
occlusion citations in Section~\ref{sec:failure}: a tracker asked to keep thirteen simultaneous,
mutually overlapping candidates straight, several of them in identical team uniforms, is working
under exactly the conditions the crowded-scene tracking literature identifies as the leading cause
of identity switches \citep{bashar2022mot,scott2024teamtrack}.}
\label{fig:overlapvolleyball}
\end{figure*}

\begin{figure}[t]
\centering
\includegraphics[width=\columnwidth]{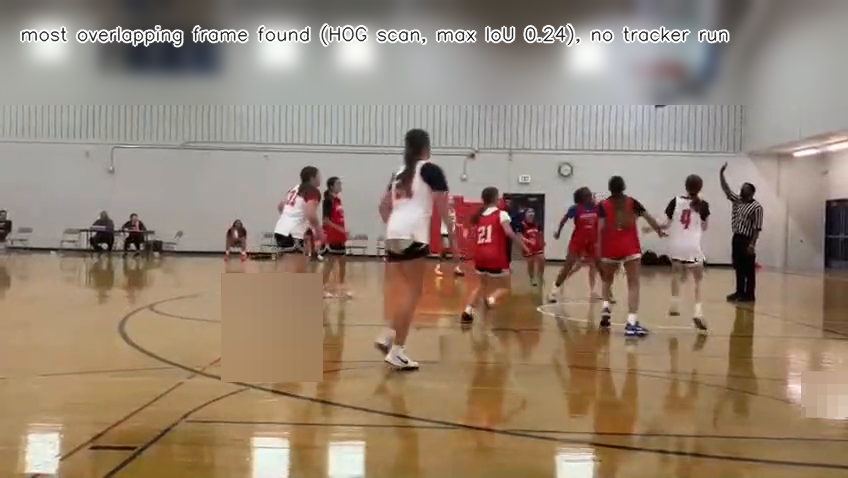}
\caption{The same crowding and occlusion pattern in a second sport, from unrelated footage no
tracker was ever run on: several players and an official occupy overlapping space at once. We
include it to show that the overlap problem in Figure~\ref{fig:overlapvolleyball} is not an
artefact of volleyball's own court geometry; Section~\ref{sec:discussion} discusses how much of it
recurs, and how much does not, as the sport changes.}
\label{fig:overlapbasketball}
\end{figure}

\begin{figure*}[t]
\centering
\includegraphics[width=0.98\textwidth]{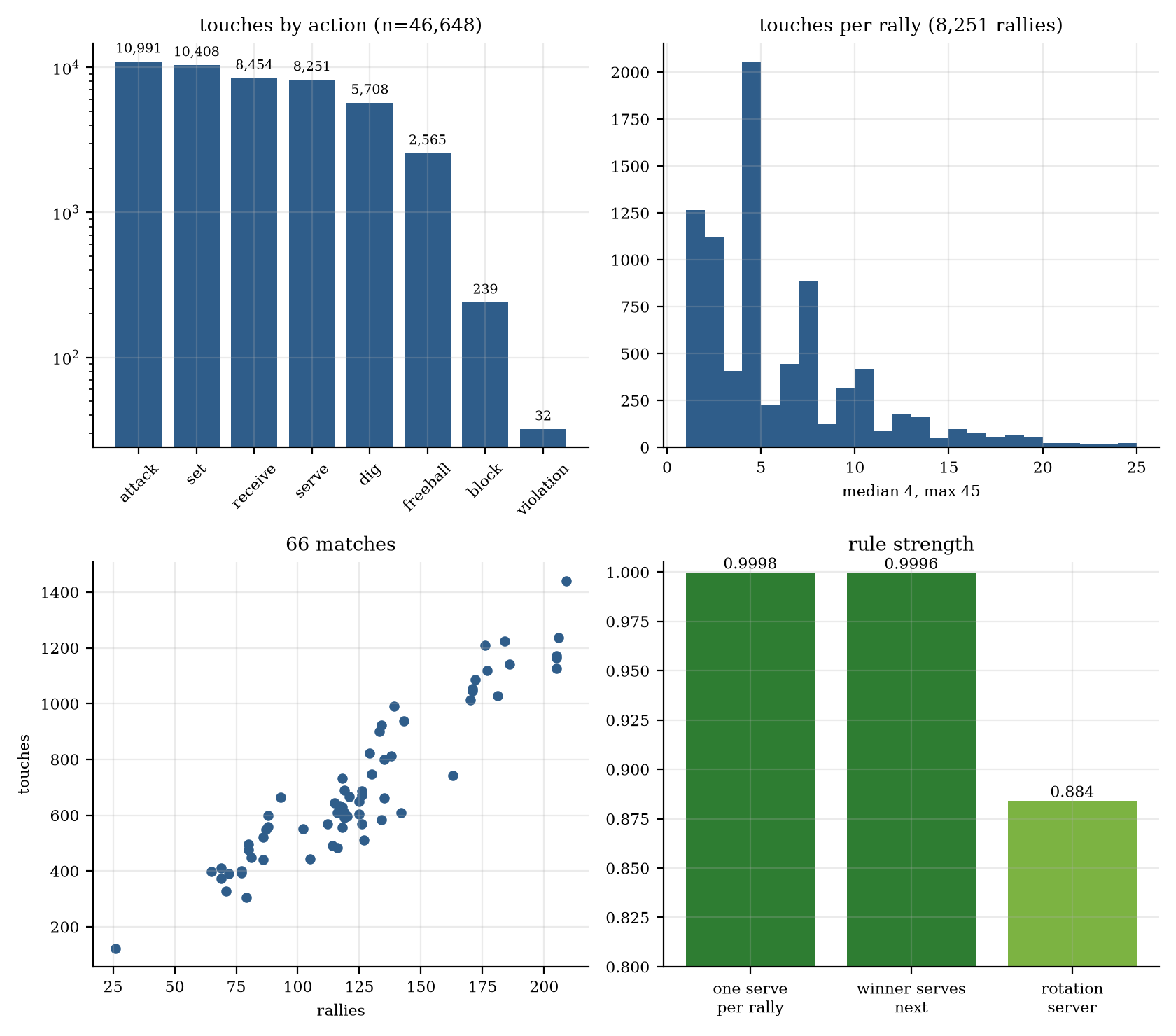}
\caption{Corpus composition. Left to right: contacts by action type (log scale); the distribution
of contacts per play; the relationship between plays and contacts across all matches in the
corpus; and the measured strength of the two rule-based invariants used throughout
Section~\ref{sec:results}.}
\label{fig:corpus}
\end{figure*}

\begin{figure}[t]
\centering
\includegraphics[width=\columnwidth]{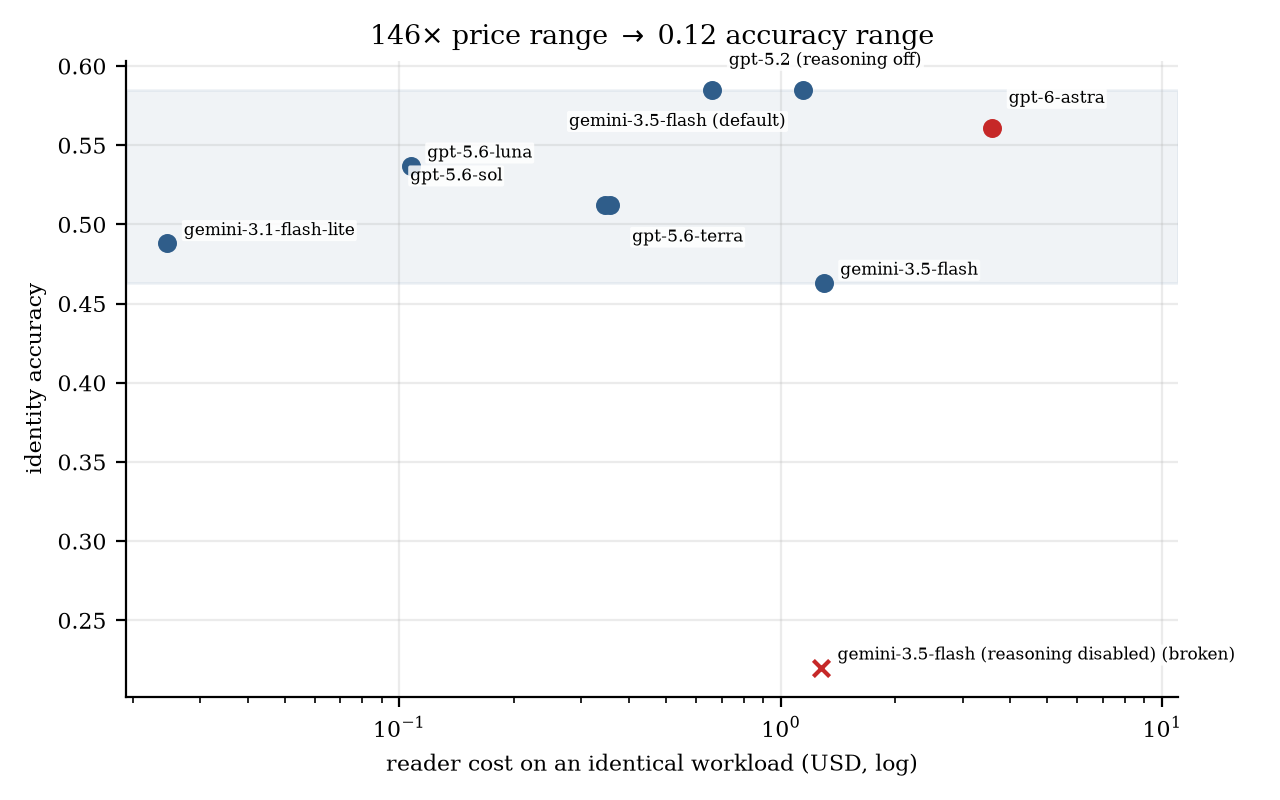}
\caption{Left: identity accuracy against metered cost for nine reading models evaluated on an
identical workload; the reader evaluated with reasoning explicitly disabled is marked separately
and excluded from the reported price-to-accuracy range because it represents a broken
configuration rather than a genuine cost/accuracy trade-off. Right: full-match cost for the
prompted, agentic, and composed-specialist paradigms on identical footage.}
\label{fig:readercost}
\end{figure}

\begin{figure*}[t]
\centering
\includegraphics[width=0.98\textwidth]{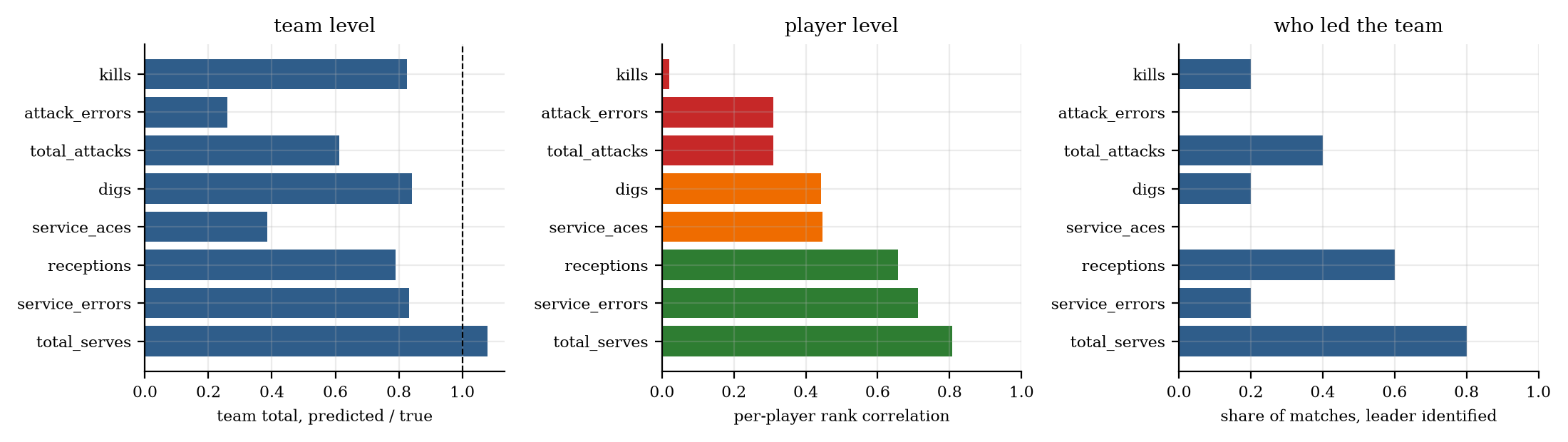}
\caption{Accuracy by statistic at three levels of aggregation: team totals (left), per-player rank
correlation (centre), and how often the correct leading player was identified (right). A
statistic's position in this figure is well predicted by how much of it the sport's own rules
already determine, as discussed in Section~\ref{sec:results}.}
\label{fig:statlevels}
\end{figure*}

\begin{figure}[t]
\centering
\includegraphics[width=\columnwidth]{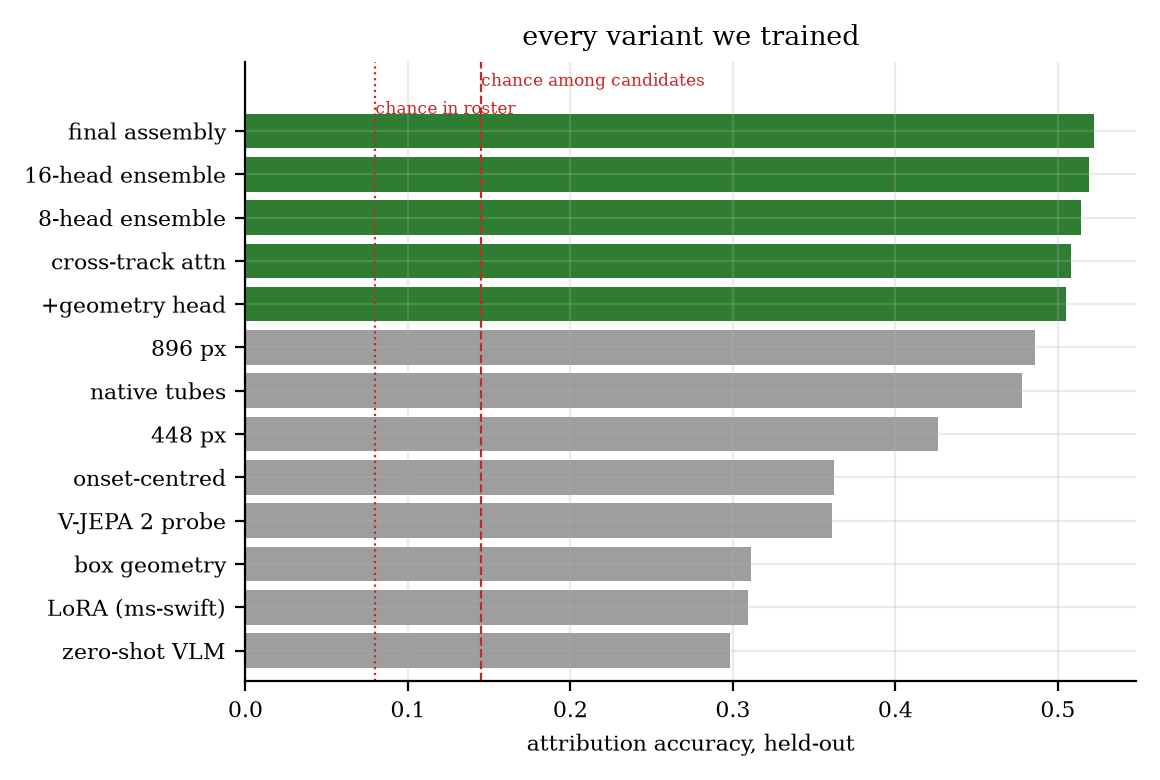}
\caption{Every selector variant we trained or probed, ordered by held-out accuracy, against two
chance baselines (random selection within the full roster, and random selection among the
candidates physically present at a contact). Only two changes moved accuracy meaningfully:
increasing working resolution, and ensembling multiple independently trained heads.}
\label{fig:trainingladder}
\end{figure}

\begin{figure*}[t]
\centering
\includegraphics[width=0.98\textwidth]{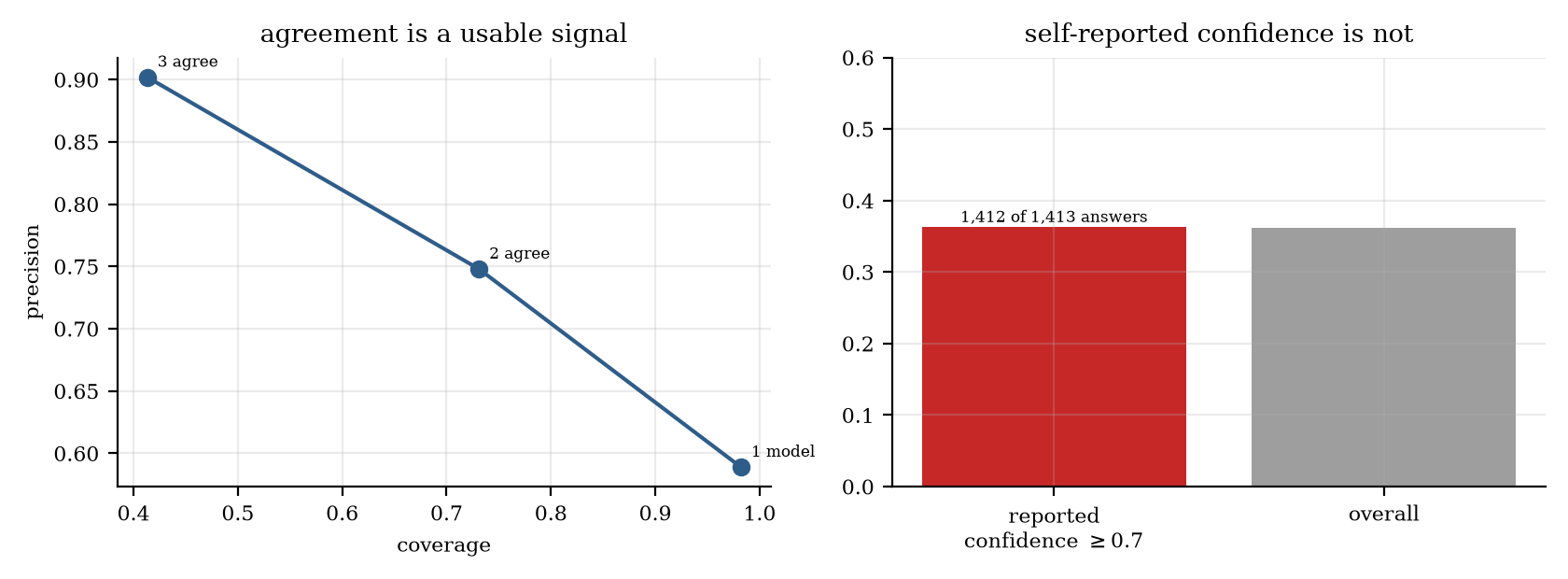}
\caption{Left: precision against coverage when abstention is driven by agreement among
independent reading passes. Right: accuracy conditional on a model's own self-reported
confidence, compared against overall accuracy; the two are statistically indistinguishable,
showing that self-reported confidence carries no useful information for abstention in this
setting.}
\label{fig:abstention}
\end{figure*}

\begin{figure}[t]
\centering
\includegraphics[width=\columnwidth]{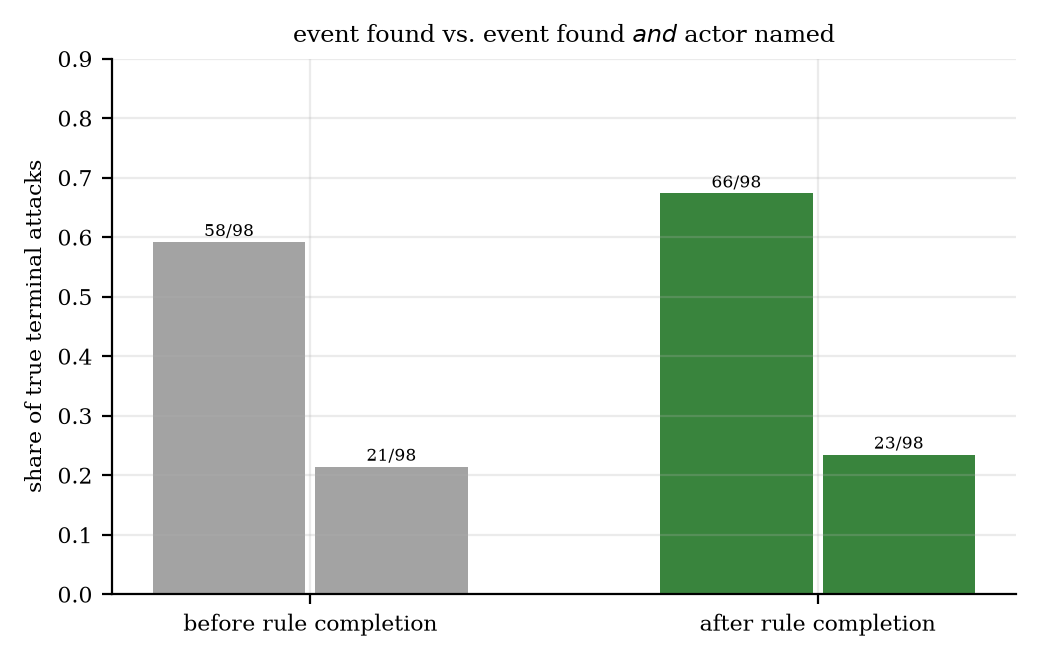}
\caption{Effect of a rule that inserts a plausible terminal attack when a play's last recorded
action is not itself an attack. Left bar pair: share of true terminal attacks for which an event
was found, before and after the rule. Right bar pair: share for which the event was found
\emph{and} the acting player correctly named. The rule substantially improves event recall but
not player attribution, the double limit discussed in Section~\ref{sec:results}.}
\label{fig:killlimit}
\end{figure}

\begin{figure}[t]
\centering
\includegraphics[width=\columnwidth]{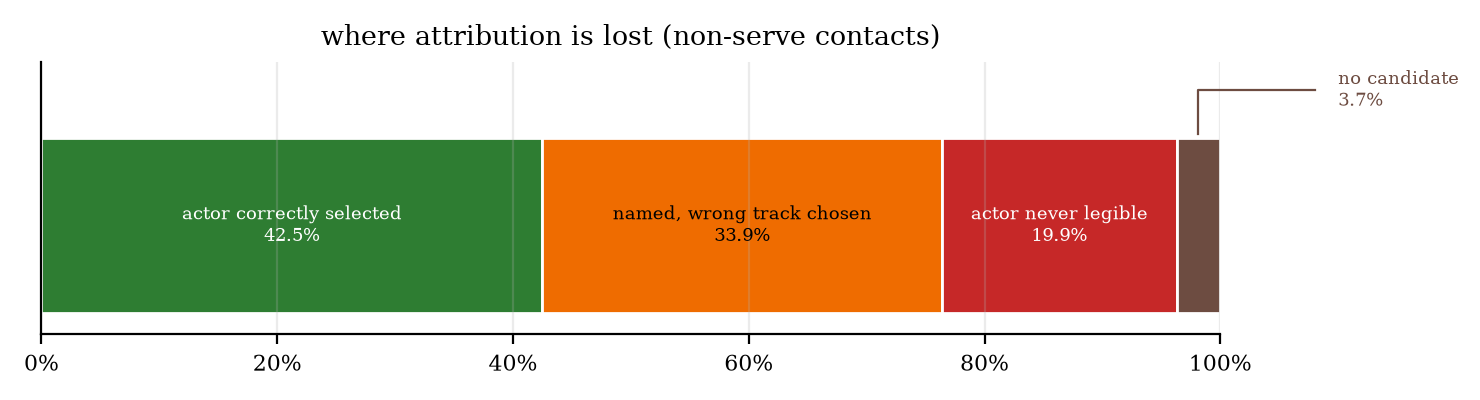}
\caption{Decomposition of attribution failure on non-serve contacts into its point of loss:
correct selection, a named candidate incorrectly passed over, the acting player never legibly
read anywhere in the play, and no viable candidate at all.}
\label{fig:decomposition}
\end{figure}

\begin{figure}[t]
\centering
\includegraphics[width=\columnwidth]{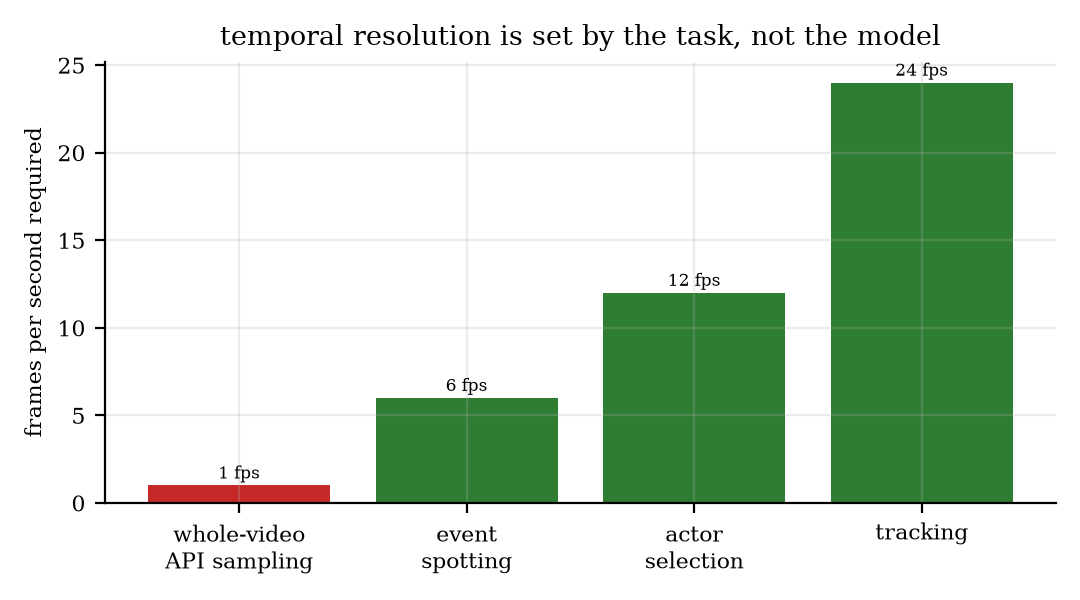}
\caption{The temporal sampling rate current commercial video-understanding APIs allot to
whole-video inference, against the sampling rate each stage in Section~\ref{sec:stages} actually
requires. Coarse segmentation tolerates the API's fixed low rate; fine-grained selection and
tracking do not.}
\label{fig:fpsbytask}
\end{figure}

\begin{figure*}[t]
\centering
\includegraphics[width=0.98\textwidth]{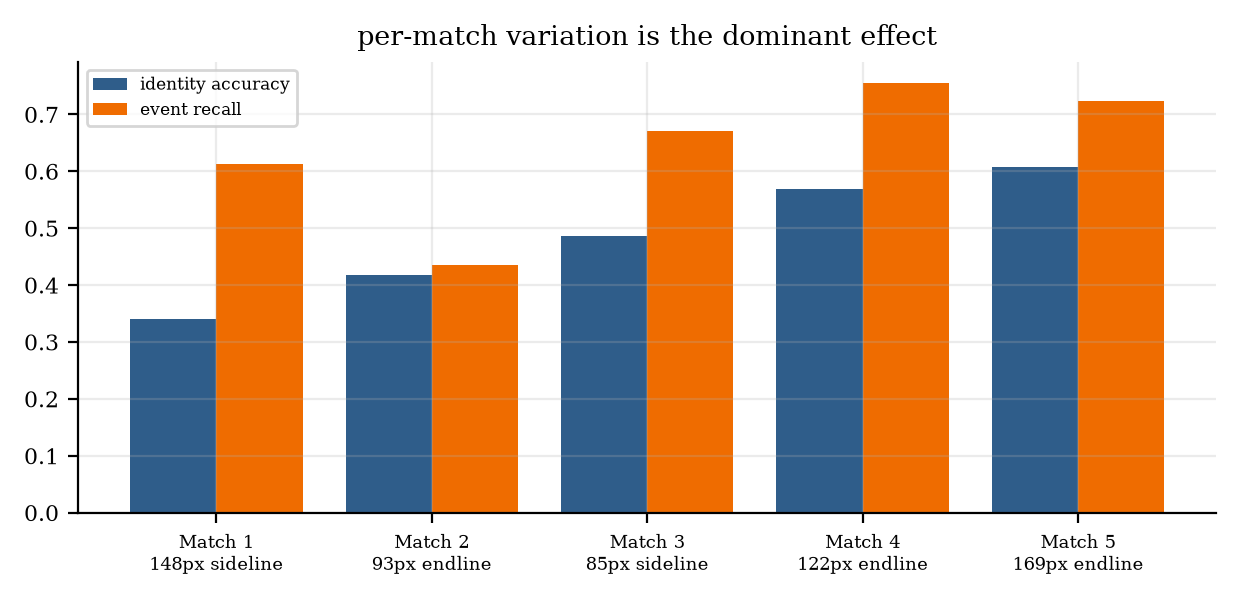}
\caption{Identity accuracy and event recall across six held-out matches, ordered by identity
accuracy, with median actor height and camera axis labelled per match. Per-match variation is
larger than the variation attributable to any single modelling change reported in
Figure~\ref{fig:trainingladder}.}
\label{fig:pergame}
\end{figure*}

\begin{figure}[t]
\centering
\includegraphics[width=\columnwidth]{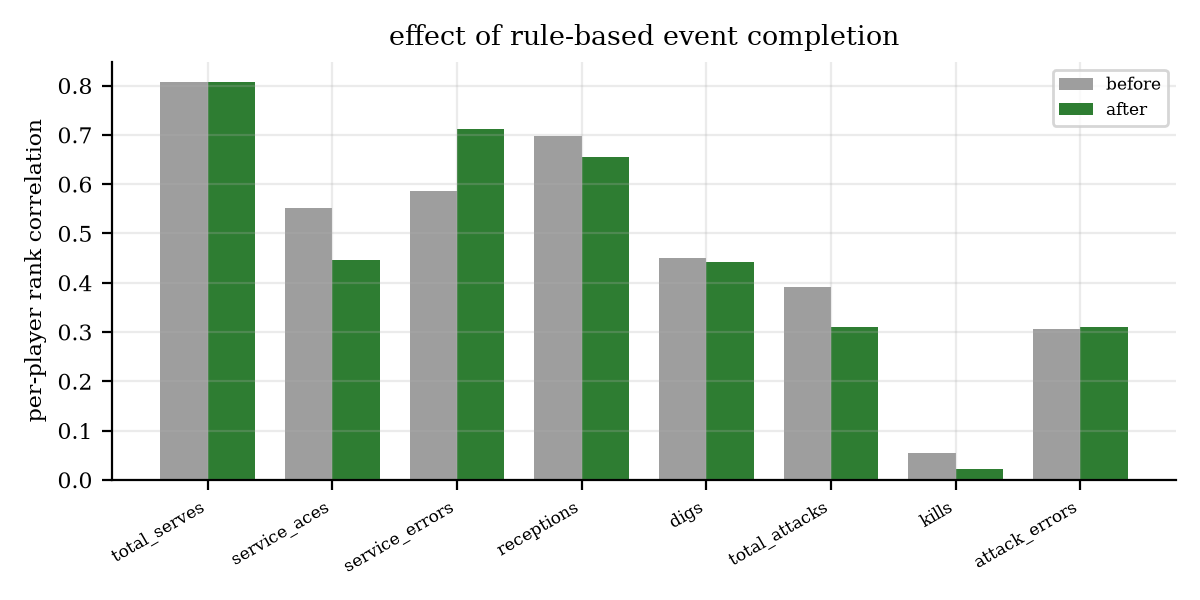}
\caption{Per-statistic effect of the terminal-event completion rule on player-level rank
correlation, before and after.}
\label{fig:ablation}
\end{figure}

\begin{figure}[t]
\centering
\includegraphics[width=\columnwidth]{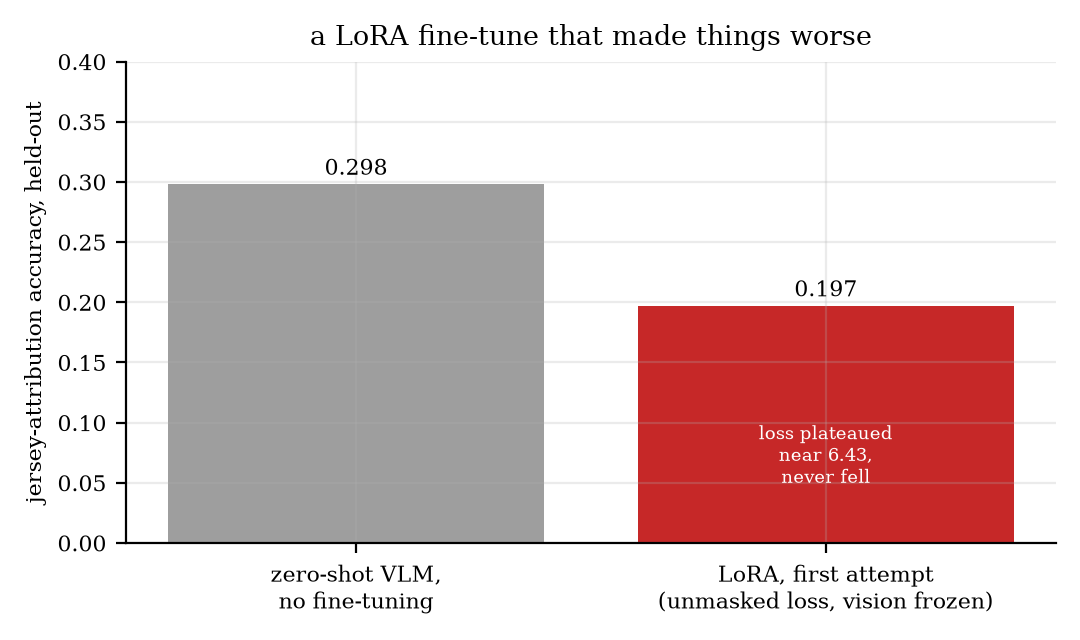}
\caption{A documented LoRA fine-tuning failure, not a hypothetical one: an early configuration
trained its loss over the whole input sequence rather than only the answer tokens, and adapted
only the language model's attention projections, leaving the vision tower frozen. Training loss
plateaued near 6.43 and never fell; the resulting adapter scored below the untrained zero-shot
baseline it was meant to improve on. Section~\ref{sec:results} discusses what fixing both faults
required and why the corrected reader still falls well short of the trained-specialist paradigm.}
\label{fig:lorafailure}
\end{figure}

\begin{figure*}[t]
\centering
\includegraphics[width=0.8\textwidth]{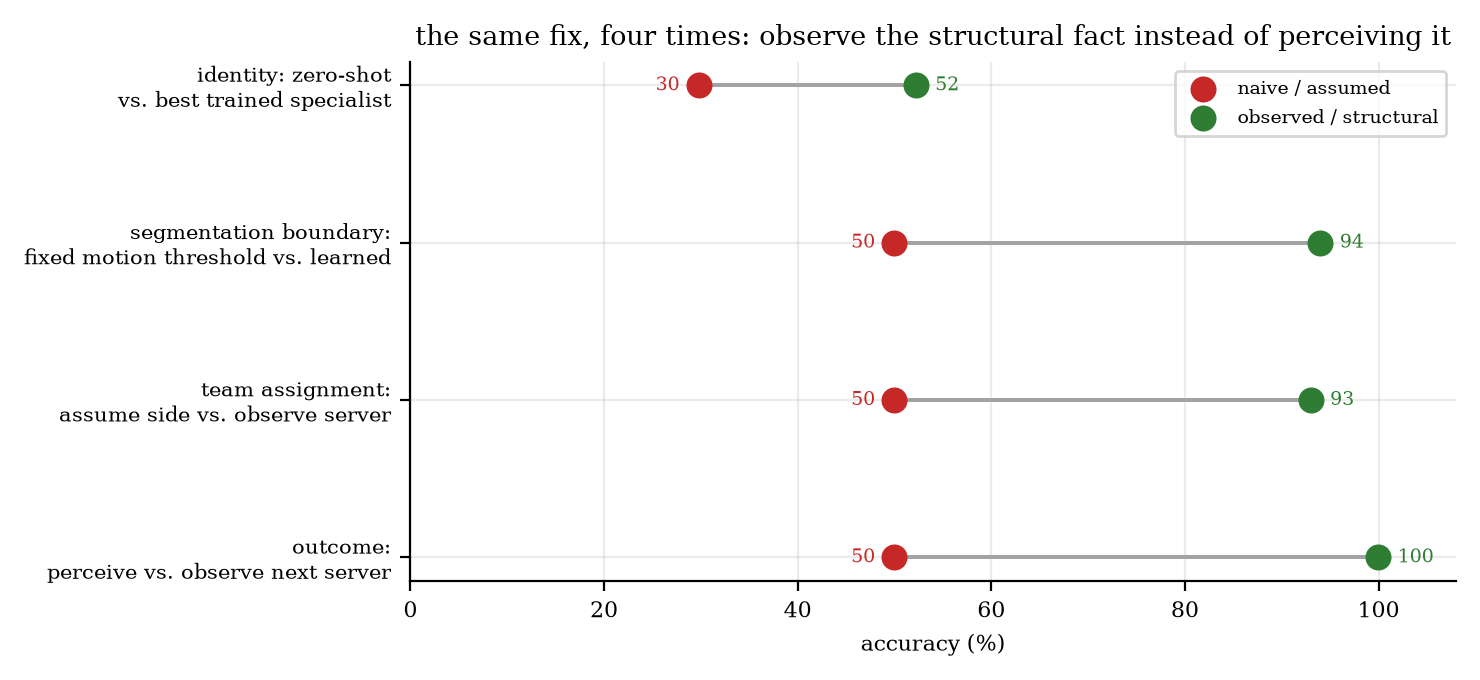}
\caption{The same pattern recurring at four different stages of this paper's pipeline: a naive,
purely perceptual or assumption-based approach (red) sits near chance, and replacing perception
with a direct observation of a structural fact the sport already fixes (green) moves it close to
ceiling. Identity is the one stage without a comparable structural shortcut, which is why it
remains this paper's central open problem rather than a fourth instance of the same fix. Every
number here is reported elsewhere in the paper; this figure only places them on one shared axis.}
\label{fig:journey}
\end{figure*}

\section{Results}
\label{sec:results}

We report results stage by stage, in the order defined in Section~\ref{sec:stages}, because the
central finding of this paper is precisely that no single ranking of the four paradigms holds
across stages. Every number below is measured on the corpus described in Section~\ref{sec:setting},
on a fixed evaluation protocol we describe at the end of this section, and every accuracy figure is
reported against per-touch human labels rather than against any single method's own output.

\subsection{Segmentation and event spotting: why this has to be temporal}
\label{ssec:segmentation}

A play boundary (the answer to ``this is a rally, from here to here'') is not a property of any
single frame. A frame taken from the middle of a live exchange and a frame taken from the dead time
immediately before the next serve can look almost identical: players standing, a ball somewhere in
the frame, no visual marker distinguishing one from the other. What distinguishes them is a pattern
of motion and event density unfolding over several seconds (the serve motion, the accelerating
exchange of contacts, the point ending, players resetting), and that pattern is only observable by
integrating information across a temporal window, never by classifying an instant. This is worth
stating explicitly because it is the first of three qualitatively different relationships between a
stage and time that recur across this paper's seven stages, and confusing one for another is
exactly the mistake that led an earlier iteration of this project to waste real engineering effort,
described below.

Because a play boundary is a genuinely temporal pattern rather than a static fact, we also tested
whether it could be recovered by simple, deterministic signal processing over the raw
video (``pure math,'' in the sense of a fixed, hand-specified rule rather than a learned temporal
model), and found that it reliably cannot. Frame-differencing to detect a scoreboard change, an
appealing low-cost proxy for a completed point, breaks on the flicker characteristic of an
LED scoreboard display, which produces frame-to-frame pixel differences indistinguishable from an
actual score update. Thresholding raw motion energy to distinguish a live rally from dead time fails
for a related reason: players resetting position, walking toward the service line, or a
referee's signal all produce motion of a comparable magnitude to live play, so no fixed threshold
generalises across venues, camera positions, or even different moments of the same match. Using the
crowd or whistle audio track as a boundary signal recovers only a partial and unreliable share of
true boundaries in our own testing, well short of what a usable segmenter requires, and multiple
concurrent matches sharing one gymnasium's acoustics make the signal noisier still. Figure~\ref{fig:motionaudio}
shows both signals computed directly on a real 40-second span of our own footage, rather than
asserted in prose: the best fixed motion threshold this specific window admits still misclassifies
roughly half of live-versus-dead time, because dead time is, at points, more motion-heavy than the
rally itself, and the audio track's clearest peak coincides with the rally's true end while
comparable peaks recur throughout the surrounding dead time. Figure~\ref{fig:audiobyaction} shows
why this is not specific to one window: measured across every one of the 518 labelled contacts in
this match, audio onset strength depends heavily on which action produced it. Hard-driven contacts
(attacks, blocks) clear a usable onset threshold a little over half the time; the sport's softer,
more defensive touches, a dig in particular, clear it barely a quarter of the time, and those
softer touches are exactly the ones that dominate a match's contact volume (Table~\ref{tab:corpus}).
An audio-based boundary or event signal is therefore not simply noisy in some fixed, uniform way;
it is systematically least informative for the majority of what actually happens in a match. We
report one
further finding against ourselves here, because it is directly instructive: earlier in this
project, a sensitivity analysis correctly identified missed play boundaries as the single most
costly class of error a system could make, and we responded by building two hand-engineered
replacements for the existing segmenter along exactly the ``pure math'' lines described
above, before confirming that the segmenter we already had was performing near the ceiling this
task admits. The lesson generalises beyond our own mistake: a sensitivity analysis correctly
identifies \emph{how costly} an error class is, not \emph{which component} is responsible for it,
and for a genuinely temporal task like this one, no fixed rule substitutes for a model that has
learned the shape of the pattern rather than been handed a threshold for it.

Given that a temporal model is required, prompting a frontier model performs about as well as any
method we evaluated at this specific stage, matching human-annotated play boundaries closely across
the corpus, because a sufficiently capable model reasoning over enough context can learn exactly the
kind of motion-and-density pattern described above without being told what to look for. A small
model trained specifically to detect play boundaries from motion and appearance features, with no
access to language or world knowledge at all, performs substantially worse at this same task,
missing or merging enough boundaries that we do not recommend it as a replacement. This is the one
stage in our study where prompting a general model is unambiguously the right choice; the
remainder of this section reports the opposite conclusion for almost every other stage.

That reversal begins immediately at the next stage down, for a precise reason: spotting an
individual contact is, if anything, an even more sharply temporal task than
segmentation: a contact is a brief kinematic event, an arm's acceleration, a ball's sudden change
of direction, embedded in continuous motion, and, like segmentation, it cannot be reliably read
from a single frame. The paradigm that wins here nonetheless changes, and the reason is not that the
task becomes less temporal but that the \emph{scale} of temporal pattern required shrinks and
sharpens: segmentation asks a system to recognise a coarse pattern unfolding over many seconds of a
whole play, a scale at which a general reasoning model's broad context is an asset; spotting asks it
to recognise a specific, brief kinematic signature at a scale closer to a fraction of a second,
where a small model built specifically to integrate motion features across a short, densely sampled
window, exactly the kind of temporal model described in Section~\ref{ssec:musclememory}, has a
structural advantage a general-purpose model sampling video coarsely does not. A compact model
trained specifically on this corpus's own labelled contacts outperforms the same frontier prompting
approach that won at segmentation, while costing on the order of a hundredth of a cent to run per
match rather than several dollars. Figure~\ref{fig:trainingladder} places this result alongside
every other selector variant we trained, ordered by accuracy against two chance baselines; the
trained spotter sits well above both, and well above the prompted alternative, at negligible
marginal cost.

\subsection{Outcome, without seeing the ball}

Determining how a play ended is, at first glance, a perception task: something happens to the ball,
and a system need only see it. In practice it is closer to the opposite. The ball in our footage
regularly measures under thirty pixels across, and a frontier model asked to classify how a rally
ended defaults overwhelmingly to the single most common outcome, misclassifying the great majority
of plays that ended any other way. Yet the correct outcome of a rally can be recovered almost
perfectly without looking at the ball at all, because volleyball's own rules determine it: the team
that serves the next rally is, with vanishing exception, the team that won the last one.
Table~\ref{tab:corpus} reports this directly: across every rally-to-rally transition in the
corpus, the team serving next is the team that won the previous rally in essentially every instance
we observed, and every rally in the corpus begins with exactly one serve. A method that simply
observes who is serving next, rather than attempting to perceive how the previous rally ended,
recovers the outcome for free. This is the clearest instance in our results of a general pattern:
where a sport's own rules fully determine an answer, no perceptual method is needed, and every
perceptual method we tried performed worse than simply consulting the rule.

\subsection{Team assignment}

Assigning a detected contact to the correct one of the two teams on court sounds, and in principle
is, simpler than naming an individual player. It nonetheless carries a specific failure mode worth
describing precisely, because it recurs, in a different guise, in almost every stage that follows
it. A system that assumes which physical side of the court belongs to which team (rather than
observing it directly for each match) will occasionally be wrong, and when it is, every contact
in that match is credited to the wrong side simultaneously. Because volleyball is scored so that
each team's statistics are summed independently, this specific error is close to invisible in
aggregate totals: a kill wrongly credited to the away team is, from the point of view of the
combined two-team total, indistinguishable from a kill correctly credited to it. It is only visible
once a system is asked to report which \emph{team}, specifically, produced a given statistic, at
which point accuracy collapses toward chance even though every downstream number appears
plausible. We measure this directly: observing which team is serving from where the server
physically stands, rather than assuming it from convention, raises team assignment from a figure
indistinguishable from a coin flip to accuracy in the low-to-mid nineties across held-out matches,
using no additional model capability at all, only a different choice of
what to observe versus what to assume. Figure~\ref{fig:teamattribution} (left) shows this directly
on a real serve rather than as a description of the mechanism.

\subsection{Identity: where every automated method converges}

Naming the individual player who made a given contact is, by a wide margin, the hardest of the
seven stages, and it is here that the comparison across paradigms is most informative.
Figure~\ref{fig:teamattribution} (right) shows what this looks like on one genuine contested
attack, where the selection step split its confidence across two spatially close, real candidate
tracks rather than committing to either. Table~\ref{tab:readers} reports jersey-reading accuracy for nine commercially available reading
models, evaluated on an identical set of collages so that only the reader itself varies. The spread
in list price across these nine readers exceeds two orders of magnitude; the spread in accuracy is
roughly ten percentage points, and the most expensive reader is not the most accurate one.
Figure~\ref{fig:readercost} plots this relationship directly. This is not, we argue, evidence that
frontier models are poorly engineered for this task; it is evidence that the information a reading
model would need, the shape of a digit occupying on the order of twenty pixels, is not reliably
present in the input at all, so additional model capacity has nothing further to extract. A reader
we evaluated with its reasoning capability explicitly disabled illustrates the same point from the
opposite direction: removing reasoning did not merely reduce accuracy modestly, it broke the reader
outright, dropping both coverage and accuracy sharply, which shows that reasoning was doing real
work compensating for genuinely ambiguous input, not padding an otherwise easy task.

Fine-tuning a vision-language model specifically on this corpus's own labelled contacts does not
close the gap. Figure~\ref{fig:trainingladder} includes this result alongside every other variant
we trained: a properly fine-tuned reader converges to an accuracy well below the strongest
trained-specialist result, and, more importantly, almost never declines to answer even on inputs
where no correct answer is visually recoverable. A system that is wrong with high apparent
confidence is worse, in a box score a coach will trust, than a system that honestly reports it
cannot tell; fine-tuning shifted what the model guessed without meaningfully improving whether it
could see. We interpret this as evidence that at the resolutions involved, fine-tuning is adjusting
the model's prior over which jersey number is likely, given the roster, rather than teaching it to
read a digit that is not legibly present.

Reaching that properly fine-tuned reader was not the first result we got, and we report the
earlier failure because it is a specific, avoidable mistake rather than a vague warning.
Figure~\ref{fig:lorafailure} shows it directly: an initial fine-tuning configuration computed its
training loss over the entire input sequence, prompt text and every image token, rather than only
the few tokens of the actual answer, and targeted adapters only at the language model's attention
projections, leaving the vision tower entirely frozen. Loss plateaued near 6.43 and never fell
across training, and the resulting adapter scored \emph{below} the untrained zero-shot baseline on
an identical held-out benchmark, 0.197 against 0.298, while abstaining on nothing. Both faults
point the same direction: a model whose vision encoder never adapts cannot improve at a task whose
bottleneck is perceptual, and a loss dominated by reconstructing thousands of prompt and image
tokens has almost no gradient left for the ten-token answer that actually matters. Masking the loss
to the answer span alone and targeting adapters at both towers, not just the language model, is
what produced the reader discussed above; we show the failed configuration alongside it because a
LoRA fine-tune that looks reasonable on paper can converge to a number worse than doing nothing at
all, and the specific reason is checkable, not mysterious.

The frozen video-world-model probe described in Section~\ref{sec:paradigms} tells a closely related
story from a third direction. A representation trained at very large scale on unlabelled video,
with a lightweight classifier fit on top and no fine-tuning of the backbone at all, achieves
identity accuracy only narrowly above a baseline constructed from nothing but the raw coordinates
of each candidate's bounding box, no pixels, no learned features, only where each candidate was
standing. This result replicates, on volleyball-specific footage, the conclusion of an earlier,
narrower comparison of the same two representation families on sports action data
\citep{kodathala2025temporal}, and it is consistent with the broader finding in the video
world-model literature that these representations, whatever their strength at coarser tasks such as
action classification, retain only fragmented, imprecise local structure rather than the exact
spatial detail fine-grained identity resolution would require \citep{murlabadia2026vjepa21}.

Composing classical detection and tracking with small, task-specific specialists is the approach
that performs best at this stage, though its best result still leaves a substantial share of
contacts unresolved. The single largest, and by far cheapest, improvement we found was resolution:
doubling the working resolution of the video clips a selector model was trained on produced a
larger accuracy gain than any architectural change we subsequently tried, including deeper
temporal models, attention-based heads, ensembles of multiple trained heads, and models trained
directly on native-resolution crops of each individual candidate rather than a shared wide view.
Figure~\ref{fig:trainingladder} reports the full ladder of roughly a dozen variants we trained in
pursuit of further gains beyond that one resolution change; almost none of them moved accuracy
meaningfully, which we report in full because a clearly reported null result is exactly the
evidence a reader needs to avoid spending a comparable amount of engineering effort re-discovering
the same ceiling.

\subsection{Whole-match reasoning: where a frontier model actually helps}
\label{ssec:wholematch}

The results above compare paradigms stage by stage, on short, bounded inputs. We report separately
what happens when a frontier reasoning model is instead given the harder and more
realistic job of reasoning over a match holistically, long spans of full-resolution footage, with
the model permitted to look, re-examine, and reconsider rather than answer from a single fixed
prompt. We evaluated this configuration using an agentic version of a capable general-purpose
Gemini model, prompted to reason step by step, to re-examine ambiguous moments, and to maintain an
explicit running account of match state as it worked through a set. On the specific stage this
configuration was built for, detecting that a scoring event of some kind occurred within a span of
play, holistic, full-resolution, agentic reasoning produced a substantial improvement over the
same family of model given only short, low-resolution clips and a single pass: team-level event
detection rose from roughly a quarter of events found, in the constrained single-pass
configuration, to a little under half in the holistic agentic one. This is a genuine and
substantial gain, and it reverses what a narrower experiment on the same models would have
suggested about the ceiling of prompting-based event detection.

That gain, however, did not carry over to identity. The same underlying model family, evaluated as
a jersey reader on identical collages under several distinct configurations, its default reasoning
setting, reasoning explicitly disabled, and a smaller, cheaper sibling model, spanned a range of
identity accuracy from a low point in the low twenties, when reasoning was disabled and the reader's
coverage collapsed with it, up to a figure in the high fifties with reasoning enabled
(Table~\ref{tab:readers}). No configuration of this model family we tested, at any price point,
closed the gap to what a coach would consider a reliable per-player read, and a materially more
expensive reader in the same evaluation was not materially more accurate. The scope of this claim
is specific: we evaluated specific, dated releases of this model family, an
agentic configuration of a capable reasoning-enabled Gemini model for holistic match reasoning, and
several flash-tier variants for jersey reading, not every subsequent point release, and API-served
models are documented to drift in capability across releases in ways not always announced in
advance \citep{chen2023howischatgptbehavior}. We would expect a materially larger or more capable version of the
same family to plausibly improve further on the event-detection side of this comparison, for the
reasons given in Section~\ref{ssec:musclememory} below; we would not expect it to close the identity
gap by a comparable margin, because the identity gap is bounded by pixels available in the source
footage, not by the reasoning capacity applied to those pixels, and no model, regardless of scale or
release date, can read information that was never optically present in the frame.

\subsection{Why context helps action, and does not help identity}
\label{ssec:musclememory}

The split described above, where holistic reasoning substantially improved event detection while
leaving identity essentially untouched, is not a coincidence of one model family, and it points to
a distinction we believe is under-appreciated in how sports video is usually discussed alongside
general-purpose video understanding. Most of what a video-understanding benchmark asks a model to
recognise is defined by appearance and context that is, in principle, recoverable from a single
well-chosen frame: an object, a scene, a static relationship between things. Sporting action is
different in a specific, mechanical way. A serve, a dig, a set, and an attack are not appearances;
they are trained, repeated motor sequences, muscle memory in the athlete's own vocabulary, and an
athlete's execution of one is, by design and years of practice, close to biomechanically invariant
across repetitions and, for the coarse-grained action category rather than fine style, close to
invariant across different athletes as well. This is a familiar experience outside sport entirely:
a person can recognise a friend from their walk alone at a distance too far to make out a face, or
recognise a familiar voice from a single laugh cut off after half a second, because a gait or a
laugh is also a stereotyped, repeated motor pattern, and the identifying information is smeared
across a short span of time rather than sitting in any one instant of it. Naming who is walking, by
contrast, still needs the face; the gait alone narrows it down without ever quite closing the gap,
which is the same shape of result this section reports for a spike and a jersey number. This is
precisely why a temporal signal, aggregated
across a short window rather than read from a single instant, carries so much more information
about \emph{what} a player is doing than any single frame does, and it is precisely why our
smallest, cheapest trained models, a temporal model reasoning over frozen appearance and motion
features with no access to language or world knowledge at all, were able to determine relative
court side and make rally outcome recoverable at accuracy no frontier model prompted directly
could match, both discussed above: the biomechanical signature of a spike is present, redundantly,
across every frame of the jump, and a model built to integrate motion over time can exploit that
redundancy directly, which is the same reason it also classifies which action a given contact was
more reliably than a model reasoning from a single instant.

Identity carries none of this redundancy, because it is not a temporal property of the action at
all. A jersey number is a static fact about a garment, printed once and unchanging for the
duration of a match; nothing about the motor sequence of a spike encodes which specific athlete is
performing it, and no amount of additional temporal context, in principle, can extract information
about a digit's shape that is not optically resolvable in at least one available frame. This is the
precise, mechanical reason context helps one stage of our pipeline and not the other, and it
explains a result we reported without full explanation in Section~\ref{sec:paradigms}: when we
measured identity visibility across a play rather than only at the instant of contact, legibility
rose sharply, roughly quadrupling for attacks and rising close to certainty for sets, not because
temporal context added information about the digit that a single frame lacked, but because a wider
temporal window gave the system more \emph{chances} to catch the one or two frames, out of many,
where the player happened to be turned toward the camera and the number happened to be
unoccluded. Context, in other words, helps identity only in its role as a \emph{search} over many
candidate moments for the rare legible one; it does not help identity in the way it helps action, by
integrating a genuinely temporal signal that is redundantly present throughout the window. Treating
these as the same kind of ``more context helps'' effect, as a first glance at our results might
suggest, would be a mistake with real consequences for where a practitioner should invest further
effort: more temporal context, more reasoning steps, and a more capable frontier model are all
reasonable investments for the action side of this problem, because the information genuinely lives
in the pattern across time; none of them, on the evidence in this paper, are a reasonable investment
for the identity side, because the information the problem needs was never distributed across time
to begin with, and either lives in a specific handful of frames or does not exist in the footage at
all.

It follows from this that the lever worth pulling was never training on more video or extracting
more features from it; every experiment in Figure~\ref{fig:trainingladder} that simply added more
of either, more parameters, more candidate frames, more independently trained heads averaged
together, produced a smaller improvement than a single change to what the model was actually given
to look at. What moved the result was a question closer to what a human coach is implicitly asking
when they watch the same footage: not how much of the clip they have seen, but whether what they
are looking at, at that moment, is actually informative about the thing they are trying to
determine. A coach recognises a spike from a fraction of a second of an arm's trajectory without
needing to see the player's face at all, and would not expect a clearer jersey number to change that
judgement; the same coach, asked who the spiker was, is doing a categorically different kind of
looking, hunting for the one moment the player turned toward them rather than integrating anything
resembling a pattern. Our best-performing systems in this paper track that same distinction rather
than treating every frame or every feature as equally worth adding, and the null results throughout
this section are best read not as failures to scale far enough but as confirmation that, for
this task, scale was never the axis that mattered.

\subsection{Why accuracy differs so much across statistics}

The four preceding stages combine, for any single box-score statistic, into an accuracy that
depends heavily on how much of that statistic's correct answer the sport's own rules already
determine. Table~\ref{tab:statistics} and Figure~\ref{fig:statlevels} report this directly across
the eight statistics we could measure at all three levels, team totals, individual player rank
correlation, and correct identification of a team's leading player, on held-out matches. Serves
and service errors, both of which the rotation rules of volleyball substantially constrain, are
recovered with rank correlations comparable to what a coach would consider a reliable read on who
is doing what.

The gap is not a matter of degree. Serve attribution reaches a player-level rank correlation of
0.81 across 303 true serves in our held-out matches, correctly names the leading server in four of
five matches, and misses a player's true serve count by an average of a little over two; kill
attribution, on the same matches and the same pipeline, falls to a rank correlation of 0.02 across
98 true kills (Table~\ref{tab:statistics}), a number statistically indistinguishable from
attributing kills at random. The reason is structural, not a matter of one stage being modelled
better than another. A serve happens at a moment volleyball's own rotation rule has already
substantially narrowed before any pixel is examined: the six players on a team serve in a fixed,
cyclically enforced order, so which of the six is due to serve next is frequently inferable from
the score and rotation alone, and the server is, by rule, the only player permitted in the service
zone at that instant, standing still, facing the court, with no opponent or teammate occupying the
same few square feet of image. A kill is decided at the opposite extreme of all three of those
properties: it is a contested, high-velocity contact that could, in principle, be made by any of
six players converging on the same volume of space at the same instant, several of them mid-jump
and turned away from the camera, with no rule narrowing the candidate set in advance. This is the
same near-versus-far distinction that raised team assignment from chance to the low nineties
earlier in this section, applied one level deeper: the fewer candidates a rule leaves standing
before perception begins, and the less those candidates are physically competing for the same
patch of court, the more of the remaining work a system can actually do.

Digs and attacks, contested and fast, sit markedly lower. Kills sit lowest of all,
and the reason is worth stating precisely, because a team-level total alone would not reveal it: at
the team level, predicted kill totals track true totals closely, which would appear, from a
dashboard showing only team sums, to be a solved problem; at the player level, the rank correlation
between predicted and true kill counts is close to zero, meaning the total is right and the
attribution underneath it is close to noise. This is the clearest instance in our results of why we
insist on reporting every statistic at all three levels rather than a single headline number: a
team total can look entirely trustworthy while concealing an attribution layer that is not.

We diagnosed this specific case further. A rule that inserts a plausible attacking contact when a
play's last recorded action is not itself an attack, because the automated spotter described in
Section~\ref{sec:paradigms} missed it, recovers a substantial share of the missing terminal events
that would otherwise be absent from the record entirely (Figure~\ref{fig:killlimit},
Table~\ref{tab:ablation}). It does not, however, meaningfully improve player-level attribution of
those same events, because the events it recovers are, almost definitionally, exactly the ones the
spotter found ambiguous enough to miss in the first place, and an ambiguous event is disproportionately
one in which the acting player is also hard to identify. Kills are, in our current system, limited
in two independent ways: the event is frequently missed, and even when recovered by rule its actor
is frequently unresolvable, and we report both limits separately rather than allowing an improvement
in one to be read as progress on the other.

\subsection{Abstention: agreement, not confidence}

Given that identity accuracy is bounded well short of perfect across every method we tested, a
production system's practical value depends heavily on whether it can distinguish its own reliable
answers from its unreliable ones, rather than presenting every guess with equal apparent
confidence. We tested two candidate signals for this. A model's own self-reported confidence score
carries essentially no useful information in our setting: the overwhelming majority of individual
reads came back reporting high confidence, and accuracy among the high-confidence subset was
statistically indistinguishable from accuracy overall, meaning the confidence score was not
tracking correctness at all. Agreement among multiple independent readers behaves very differently.
Figure~\ref{fig:abstention} plots precision against the resulting coverage: contacts where multiple
independent reading passes agree are correct at a rate high enough to present to a coach without
qualification, while the minority of contacts where readers disagree are exactly the contacts a
system should route to a human reviewer rather than guess on. This distinction, between a model
reporting how sure it feels and a system independently checking whether it is
right, is, we believe, the single most actionable finding in this paper for anyone building a
product on top of an imperfect perception system: build abstention from disagreement between
independent estimates, not from a model's own stated confidence in a single one.

\subsection{Temporal resolution is set by the task, not the model}

A recurring practical constraint across every prompting-based result in this section is the rate at
which commercial video-understanding APIs actually sample the footage they are given, independent
of the resolution or duration of the source video. Figure~\ref{fig:fpsbytask} compares the sampling
rate current APIs allot to whole-video inference against the temporal resolution each stage in
Section~\ref{sec:stages} actually requires: coarse segmentation tolerates a low sampling rate quite
well, while the fine-grained kinematic detail player attribution depends on requires a
substantially higher one, closer to what a trained specialist model operating on explicit frame
sequences can be given directly but a general-purpose video API, sampling at a fixed low rate
regardless of what is asked of it, structurally cannot supply. This is a second, independent
explanation, alongside the resolution argument in Section~\ref{sec:paradigms}, for why prompting
performs unevenly across stages: it is not simply that a frontier model is better at some tasks
than others, but that the infrastructure delivering video to that model is tuned for a sampling
rate appropriate to coarse scene understanding, not to the frame-level precision fine-grained
attribution needs.

\subsection{A worked example: the same rally, three real requests}
\label{ssec:workedexample}

The argument above is easiest to see as a live demonstration rather than only as a claim about API
defaults, so we ran one directly and report exactly what came back, prompts and all. We took a
single real 13.33-second rally from our held-out corpus (7 human-labelled contacts: serve, receive,
set, attack, dig, set, attack, in that order) and sent it to the same reader from
Table~\ref{tab:readers} (gemini-3.5-flash) three times, changing exactly one variable each time.
The prompt for the first two requests was identical:

\begin{quote}
\small\ttfamily
This is a single continuous volleyball rally, N frames sampled uniformly across 13.3 seconds of
real match footage. List every distinct moment a player contacts the ball, in order. For each, give
an approximate timestamp in seconds from the start of this clip and the contact type: one of serve,
receive, set, attack, dig, block, freeball. Return ONLY JSON: \{"contacts": [\{"t": $<$float$>$,
"action": "$<$type$>$"\}, ...]\}
\end{quote}

At one frame per second (13 frames, the sampling rate Figure~\ref{fig:fpsbytask} shows current
whole-video APIs default to), the model's complete, well-formed response was
\texttt{\{"contacts": []\}}: zero contacts found, on a rally that contains seven. At three frames
per second on the identical rally (40 frames, still far below what our own trained spotter uses),
the same model, same prompt, found ten contacts spanning the same
serve-receive-set-attack-dig-set-attack pattern the true sequence follows, with timestamps
drifting by roughly a second by the end
of the clip and three extra events beyond the seven true ones. Frame rate alone, holding the prompt
fixed, is the difference between finding nothing and finding a recognisable, if imperfect, version
of the truth.

The third request held the frame rate fixed at three per second, the configuration that had just
found ten plausible contacts, and changed only the prompt, to the generic instruction a
general-purpose video-understanding benchmark might use: ``Describe what is happening in this
video.'' The response was fluent and largely accurate as prose (it correctly identifies the two
teams by jersey colour and describes a serve-receive-set-attack-dig sequence), but it contains no
timestamp, no discrete event count, and no structure a box score could be built from; it also
narrates a blocked point and a celebration that do not appear in the labelled ground truth for this
rally. The same forty frames, the same model, produced either a countable, if imperfect, event list
or a fluent paragraph with an invented ending, depending only on how the question was asked. Frame
rate determines whether the information needed is delivered to the model at all;
the prompt determines whether the model is asked to structure what it saw into something a
downstream system can count. Neither substitutes for the other, and general-purpose video
understanding, as it is typically benchmarked, controls for neither.

\subsection{Why accuracy is not the metric, and how a multi-event match is actually scored}
\label{ssec:metrics}

We close this section by stating our evaluation protocol precisely, because several of the results
above depend on the details of how a prediction is scored, not only on what was predicted, and
because a single accuracy number, applied to a match that produces hundreds of contacts across
eight or more distinct statistics, would actively hide the finding this paper exists to report. A
match is not one prediction to be marked right or wrong; it is a long sequence of events of several
different kinds, occurring at wildly different rates, some of which the sport's rules almost fully
determine and some of which no method we tested resolves reliably (Section~\ref{sec:results}).
Pooling all of them into one accuracy figure would let the easy, high-volume statistics, serves in
particular, mathematically dominate the number and conceal the hard, low-volume ones underneath it,
which is exactly the effect that let a team-level kill total look solved in
Section~\ref{sec:results} while the player attribution beneath it was close to noise. We therefore
never report one accuracy figure for a match. Every statistic is scored on its own, at every level
of aggregation it is reported at, and the reader is shown the full spread (Table~\ref{tab:statistics},
Figure~\ref{fig:statlevels}) rather than a single blended average standing in for it.

Within a single statistic, a system's output is itself a sequence of individual events, one
attribution per contact, not one judgement per match, and each of those events is scored
independently against its own matched ground-truth event before anything is aggregated. A contact a
system got right does not offset a contact it got wrong elsewhere in the same rally; correctness is
never averaged across events before it is counted, only after. This is also why accuracy in the
single-number sense is the wrong instrument even within one statistic: a system that is allowed to
abstain, as ours is, is not answering every event, so accuracy alone conflates three genuinely
different outcomes, a correct answer, a wrong answer, and a declined one, into two categories, and a
reader cannot tell from an accuracy figure alone whether a low score means the system is often wrong
or often silent. We therefore report coverage (the share of contacts a system actually answered, as
opposed to declining to answer) separately from precision on the answered subset, because collapsing
the two into one number obscures exactly the trade-off Section~\ref{sec:results} shows to be the
most actionable finding in the paper, and we report every tier of that abstention separately rather
than averaging across tiers, since a tier that is right 80\% of the time and a tier that is right 20\%
of the time convey materially different instructions to a coach even when their unweighted average
happens to coincide with some other system's single reported number.

We do not score against timestamps directly, because doing so would let a segmentation or spotting
error masquerade as an attribution error. Instead, each stage is scored conditional on the preceding
stage having succeeded: a predicted play is matched to a true play by temporal overlap, with a
tolerance that accounts for the corpus's one-second label resolution; a predicted contact is
matched to a true contact only if it falls within a fixed time window of the true contact \emph{and}
carries the same action label; identity is scored only on contacts that both matched successfully
at the action level and carry a jersey number in the underlying human label, which excludes plays
where the label itself does not identify an individual. For per-player statistics specifically, we
report rank correlation between predicted and true counts rather than exact-count accuracy, because
a coach's actual use of the number is comparative, who led the team, who is improving relative to
whom, rather than absolute, and a system that consistently over- or under-counts by a stable margin
while preserving the correct ordering is, for that purpose, more useful than a system with a lower
mean error but a scrambled ranking; Table~\ref{tab:statistics} reports both, precisely so a reader
can see where the two diverge. Any uncertainty in the figures above should be judged by resampling
whole matches rather than individual contacts, since contacts within one match are not independent
observations of the underlying accuracy; the per-match breakdown in Table~\ref{tab:pergame} is the
practical form this takes in this paper, showing the actual match-to-match spread directly rather
than collapsing it into a single interval around a pooled estimate. We report macro statistics
alongside volume-weighted ones throughout,
because the two diverge substantially on our data and a single number would silently favour
whichever statistic happens to occur most often in the corpus.

\begin{table}[t]
\centering
\caption{Nine readers evaluated on an identical set of jersey-reading requests: same crops, same
tracks, same prompt. Cost is the metered spend for that workload; accuracy is exact-match jersey
identification. The reader with reasoning disabled failed outright and is shown for completeness,
not included in the reported price-to-accuracy range.}
\label{tab:readers}
\begin{tabular}{lrr}
\toprule
Reader & Cost (identical workload) & Identity accuracy \\
\midrule
gemini-3.1-flash-lite & \$0.025 & 0.488 \\
gpt-5.6-luna & \$0.107 & 0.537 \\
gpt-5.6-sol & \$0.345 & 0.512 \\
gpt-5.6-terra & \$0.357 & 0.512 \\
gpt-5.2 (reasoning off) & \$0.661 & 0.585 \\
gemini-3.5-flash (default) & \$1.142 & 0.585 \\
gemini-3.5-flash (reasoning disabled) & \$1.275 & 0.220 \textit{(disabled reasoning; excluded from range)} \\
gemini-3.5-flash & \$1.296 & 0.463 \\
gpt-6-astra & \$3.585 & 0.561 \\
\bottomrule
\end{tabular}
\end{table}

\begin{table}[t]
\centering
\caption{Accuracy by statistic at three levels of aggregation, five held-out matches with
per-touch ground truth. \emph{Team ratio} is the predicted team total divided by the true team
total. \emph{Player rank} is the Spearman rank correlation between predicted and true per-player
counts. \emph{Leader} is how often the top scorer was correctly identified. \emph{MAE} is the mean
absolute per-player error. A statistic can look accurate at the team level while carrying almost
no information at the player level, which is why we report every level separately rather than a
single number.}
\label{tab:statistics}
\begin{tabular}{lrrrrr}
\toprule
Statistic & Team ratio & Player rank & Leader found & MAE/player & $n$ (true) \\
\midrule
Serves & 1.08 & 0.81 & 4/5 & 2.18 & 303 \\
Service errors & 0.83 & 0.71 & 1/5 & 0.35 & 36 \\
Receptions & 0.79 & 0.66 & 3/5 & 2.73 & 285 \\
Aces & 0.38 & 0.45 & 0/5 & 0.58 & 39 \\
Digs & 0.84 & 0.44 & 1/5 & 2.82 & 221 \\
Attacks & 0.61 & 0.31 & 2/5 & 5.69 & 439 \\
Attack errors & 0.26 & 0.31 & 0/5 & 1.14 & 73 \\
Kills & 0.83 & 0.02 & 1/5 & 1.91 & 98 \\
\bottomrule
\end{tabular}
\end{table}

\begin{table}[t]
\centering
\caption{Effect of adding a rule that completes an undetected terminal attack when a rally's
last recorded action is not itself an attack. The rule recovers missing events but the identity
of the acting player is not recovered along with it, so the effect on player-level rank
correlation is mixed rather than uniformly positive.}
\label{tab:ablation}
\begin{tabular}{lrrr}
\toprule
Statistic & Before & After & Change \\
\midrule
Serves & 0.81 & 0.81 & +0.00 \\
Service errors & 0.59 & 0.71 & +0.13 \\
Receptions & 0.70 & 0.66 & -0.04 \\
Aces & 0.55 & 0.45 & -0.11 \\
Digs & 0.45 & 0.44 & -0.01 \\
Attacks & 0.39 & 0.31 & -0.08 \\
Attack errors & 0.31 & 0.31 & +0.00 \\
Kills & 0.05 & 0.02 & -0.03 \\
\bottomrule
\end{tabular}
\end{table}

\begin{table}[t]
\centering
\caption{Per-match variation, the five held-out matches with per-player jersey ground truth (a
sixth held-out match without it is excluded from this table). Camera axis and median actor height
are measured from the tracked boxes, not from metadata. Identity accuracy does not order
monotonically with either alone; the two-way combination is the better predictor. Footage tier is
our own qualitative judgement of how usable that match's footage was overall, not a separate
quantitative metric: \emph{reliable} matches support the per-player results reported throughout
this paper without qualification, \emph{team level only} matches are usable for team-level
statistics but not trusted for per-player ones, and \emph{partial} matches have gaps in coverage
that limit both.}
\label{tab:pergame}
\begin{tabular}{lrrrl}
\toprule
Camera axis & Actor height (px) & Identity acc. & Event recall & Footage tier \\
\midrule
Endline & 169 & 0.608 & 0.724 & reliable \\
Endline & 122 & 0.569 & 0.755 & team level only \\
Sideline & 85 & 0.487 & 0.670 & partial \\
Endline & 93 & 0.418 & 0.435 & team level only \\
Sideline & 148 & 0.340 & 0.612 & partial \\
\bottomrule
\end{tabular}
\end{table}

\section{Why Identity Fails, and Why Structure Recovers Team State}
\label{sec:failure}

Section~\ref{sec:results} established that identity is the stage every paradigm struggles with, and
that team-level and outcome-level state can instead be recovered almost for free from the sport's
own rules. This section explains why both of those things are true, decomposes the identity failure
into its component causes, and connects each cause to the broader literature introduced in
Section~\ref{sec:related}.

\subsection{Where identity is lost}

We traced every non-serve contact in our held-out evaluation through the pipeline described in
Section~\ref{sec:paradigms} and recorded, for each one, the specific point at which correct
attribution was lost. Figure~\ref{fig:decomposition} reports the resulting decomposition. A small
minority of contacts have no viable candidate track at all, typically because a player was
occluded by a teammate or an official at the moment of contact, the same failure mode reported
broadly across crowded-scene tracking, where occlusion by another target is a leading cause of
detection dropouts, track fragmentation, and identity switches \citep{bashar2022mot}, and one
documented as a persistent, sport-specific problem in its own right, since players on the same team
are, by uniform, close to indistinguishable to a tracker even before any occlusion occurs
\citep{scott2024teamtrack}. Figure~\ref{fig:overlapvolleyball} shows the single worst instance of
this we could locate anywhere in our tracked corpus, found by searching every frame for the
greatest pairwise box overlap among simultaneously present candidates rather than picked by eye: it
is exactly the kind of frame that turns tracking from a bookkeeping problem into a genuinely hard
one. A substantially larger share have
candidates, but the jersey number of the player who actually acted is never legibly read on any of
them across the entire play, and the information genuinely was not recoverable from the footage, by
any method, at any point. The single largest share of losses, larger than either of the two above,
occurs where the acting player \emph{was} legibly read somewhere in the play, and was present among
the candidates offered to the selection step, but a different candidate was chosen instead. This
last category is the one that most directly motivates the resolution experiments reported in
Section~\ref{sec:results}: it is a selection failure conditional on identity already being
available, rather than an identity-reading failure, and it is precisely the category that
doubling working resolution measurably improved.

Figure~\ref{fig:tracklet} shows why the largest of these three failure categories, an identity
that was read correctly somewhere in the play but not selected at the moment it mattered, is
recoverable at all. The same tracked player, sampled across seven points of a single continuous
play, is clearly legible at some of those points and illegible at the rest, as the player turns,
moves, and is briefly screened by an opponent. A system that is only permitted to read identity at
the instant of a contact will, on this evidence, frequently choose one of the illegible moments by
construction, since a contact is itself a moment of rapid motion. A system that tracks the player
across the whole play and reads identity wherever, across that track, the player happens to be
most legible converts an unfavourable single-instant read into a favourable whole-track one, which
is the architectural choice underlying the trained-specialist paradigm's advantage at this stage.

\subsection{Why a top-down architecture outperforms end-to-end reasoning here}

The approach that performed best at identity, detecting every person first, following each one
across frames, and only then asking which followed person acted, is not the only architecture a
system could take. Why it outperformed the alternative of asking a single model to reason directly
over raw video is explained below, not merely asserted. Four properties of this
decomposition matter. First, the detection step is never asked to read anything; it only has to
localise a person, a task well within the resolution of our footage, deferring the harder reading
problem to a later, better-positioned step. Second, decomposing the problem this way converts an
open-ended question, which of an entire roster this contact could belong to, into a bounded
choice among a small number of tracked candidates physically present in the frame, which is a much
easier problem even before any model quality is brought to bear. Third, and most importantly,
tracking allows a system to read a player's identity once, at whichever moment across an entire
play that player happens to be most legible, and then propagate that reading to every other contact
belonging to the same track, rather than being forced to read identity at the specific instant of
each individual contact, which is frequently the single worst moment to attempt it because the
player is mid-motion, turned away from the camera, or partially occluded by an opponent. Fourth,
because the pipeline is decomposed into discrete stages, a failure is attributable to a specific
stage rather than presented as one opaque wrong answer, which is precisely what makes the
decomposition in Figure~\ref{fig:decomposition} possible to compute at all.

This architectural choice is supported directly by evidence from outside our own corpus. A recent
comparison of a supervised, specialised object detector (a model trained specifically to draw a box
around every instance of a chosen category, here a person, without saying anything about which
specific person it is) against general-purpose vision-language models on the same standard object
categories finds the specialised detector roughly twenty percentage points more accurate, a gap
that narrows, but does not close, on categories the detector was never trained on
\citep{alhamadani2025hiddeneconomics}. In a materially different but structurally related result,
aligning a detector's own features precisely to each candidate region, rather than pooling coarsely
across it, is itself a measurable, separable source of comparable accuracy gains in the classic
instance-segmentation literature \citep{he2017maskrcnn}. On sports footage specifically, the same
track-then-read structure we use here, with confidence-weighted voting across multiple readings of
the same tracked identity, is the approach reported to achieve the strongest published jersey-number
accuracy on comparable footage, and its own authors attribute a specific, measurable share of that
result to exactly the main-subject filtering and consolidation steps our own design mirrors
\citep{koshkina2024jersey}.

The alternative, of asking a model to look more carefully at an ambiguous region through iterative
zooming or cropping rather than through tracking, does not close the gap, and the broader
agentic-vision literature explains why with some precision. A controlled study of what a
crop-and-zoom reinforcement-learning loop actually contributes to accuracy finds that the great
majority of any measured improvement is attributable to the model's pre-existing capability rather
than to the zooming action itself, with the zoom's own contribution falling to a small single-digit
share specifically on the hardest, most fine-detail benchmarks tested \citep{ma2026visiontooluse}.
A closely related study of iterative visual self-correction for grounding finds that an apparent
accuracy gain from iterating disappears entirely once the evaluation is restricted to a stopping
rule a real system could actually deploy, because the model's own signal for whether a given
iteration has improved or worsened its answer is only weakly correlated with whether it actually
has \citep{tripathy2026selfcorrectionmirage}. Both findings match our own zero-shot and single-pass
prompting results directly: an agentic system that reasons and re-examines the same footage
repeatedly does not, in our measurements, recover information that was not present in the footage
to begin with, because the difficulty is a property of the pixels rather than of how many times a
model is permitted to look at them.

\subsection{Why global rules help and local chaining hurts}

A further finding, reported briefly in Section~\ref{sec:results} and worth explaining here, is that
constraining a system's output using the sport's rules helps substantially when the constraint is
global, a property that holds across an entire play or match regardless of any single local
decision, such as ``exactly one serve begins every rally'', and actively hurts when the constraint
is local, chaining one uncertain decision's output into the next decision's input, such as
excluding whichever player was attributed the previous contact from consideration for the current
one. Every local-chaining constraint we tried reduced accuracy relative to treating each contact's
attribution as an independent decision, because a single wrong upstream decision, once chained
forward, actively removes the correct candidate from consideration at the next step rather than
merely failing to help. This is the same failure mode documented in the multi-agent language-model
literature under a different name: sequential pipelines in which each stage trusts its
predecessor's output without independent verification show measurable, accumulating error across
successive stages \citep{singh2026hallucinationsnowball}, and a simpler pipeline with fewer chained
decisions can measurably outperform a more elaborate one for exactly this reason
\citep{prajapati2026twocalls}. Global constraints do not share this failure mode because they do not
depend on any single earlier decision being correct; they hold, or fail to hold, independent of the
order in which a system happens to process events, which is precisely why observing the game state
directly, who is serving, which team just won a rally, outperforms deriving it by chaining a
sequence of local inferences together.

\subsection{Why a fixed threshold does not travel across cameras}

A related and, in deployment, more consequential failure mode than any single-corpus result above
is that a threshold calibrated on one camera does not transfer cleanly to another, even when the
two cameras are filming the same sport from a similar distance. Section~\ref{sec:setting} already
showed that pixel height alone is an incomplete predictor of whether a jersey number is legible,
because viewing angle confounds it: an end-line camera keeps a flat, legible number readable at
heights a sideline camera cannot resolve at all. The same confound reappears one level earlier, in
the camera itself, before any player-side variation is considered. A given number of pixels across
a person's jersey number represents a different real-world size, and a different share of that
person's actual height, depending on the sensor, the lens's focal length, and the distance between
the camera and the court, none of which a fixed pixel-height threshold can see; two cameras placed
at the same physical distance from the same rally, differing only in focal length, will hand a
downstream reader two different pixel heights for the identical digit at the identical moment.
Object detection tolerates this variation reasonably well, because a silhouette remains a
silhouette across a wide range of scales. Pose estimation and fine-grained reading do not, because
both depend on resolving a specific, small structure, a joint, a digit, whose visibility is exactly
the quantity a change in focal length or distance perturbs.

This is not a claim specific to volleyball or to our own corpus. A controlled comparison of
markerless pose estimation against marker-based ground truth, using multiple synchronised cameras
on the same subjects, reports error concentrated specifically in the frames where the pose
estimator's own tracking broke down given a particular camera's view of a limb, rather than
declining smoothly with a single resolution number \citep{nakano2020openpose}, and a separate
validation of consumer action-camera rigs for the same class of motion-analysis reconstruction
reports that reconstruction accuracy is sensitive to which specific camera combination and which
calibration procedure was used, not to distance alone \citep{johnson2023actionsportcameras}. Both
findings say what our own setting-level result already says: a threshold, or an accuracy figure,
reported for one camera configuration is a statement about that configuration, not a portable
constant, and a system whose deployment behaviour is described only in terms of a pixel count or a
single reported accuracy number is under-describing the conditions that number actually depends on.

This is also not merely an academic caution. The same underlying promise, that an ordinary phone
camera's video can be converted into research-grade motion measurement, is being commercialised
well outside sport, in workplace ergonomics assessment, by vendors offering markerless,
single-camera motion capture as a product; one such vendor is, as of this writing, in active
litigation and a court-supervised insolvency process \citep{motionai2025insolvency}. We cite this
only to make a narrow point: the gap between a calibrated demonstration and a deployed product that
must work across whatever camera a customer happens to already own is a live commercial pressure in
this broader space, not a concern specific to our own corpus, and it is one reason we report every
accuracy figure in this paper against the specific camera conditions described in
Section~\ref{sec:setting} rather than as an unconditional number.

\section{Deployment}
\label{sec:deployment}

The results in Section~\ref{sec:results} were obtained not only in controlled offline experiments
but also from a production system that has processed real matches submitted by real users, and it
is worth reporting what operating that system at that scale actually looks like, without disclosing
the specific engineering choices that make it economical to run, for the reasons given in
Section~\ref{sec:statements}.

The system processes many matches concurrently rather than one at a time, and a match whose footage
has already been processed elsewhere in the system returns its result without repeating the
underlying computation. Under concurrent load, aggregate throughput is substantially higher than
serial processing of the same workload would achieve, and a single new match, processed against
otherwise available capacity, completes well within the length of the match itself. This
throughput matters in a specific, concrete way for the comparison drawn throughout this paper: the
prompting-based paradigm's per-stage costs, reported in Section~\ref{sec:results} as a metered rate
against an identical workload, translate directly into a full-match cost, summed across all seven
stages, that is several times higher than the composed system's, even though the gap at any single
stage can be far larger still (Section~\ref{sec:results} reports a difference of several orders of
magnitude at the spotting stage specifically), without a corresponding accuracy advantage at most
stages, which
is the central economic finding of this paper stated in production terms rather than experimental
ones.

The system also implements the abstention mechanism described in Section~\ref{sec:results} as a
user-facing feature rather than only an offline metric. Rather than presenting every player
attribution with equal apparent confidence, the interface distinguishes three tiers: attributions
confirmed by independent agreement and presented without qualification, attributions where readers
disagreed and a short list of the most likely candidates is offered for a human to confirm with a
single interaction, and contacts where no reliable candidate could be identified at all, routed
directly to a review queue rather than guessed at. This design follows directly from the finding in
Section~\ref{sec:results} that a model's own confidence score is not a useful abstention signal:
the tiers are built from inter-reader agreement, not from any single model's stated certainty. Every
correction a reviewer makes through this interface is also a new training signal, and confirming a
player's identity once, for a single contact, measurably improves accuracy on the remaining
unconfirmed contacts belonging to the same tracked player, since correcting one contact propagates
the corrected identity across the entire track rather than only the single event a reviewer looked
at.

We report the volume of manual review this system currently requires because it bears
directly on the comparison with human annotation drawn throughout this paper: a coach using the
system today reviews a minority of a match's total contacts, concentrated specifically in the
low-confidence tier the system itself identifies, rather than reviewing the match in its entirety
as manual annotation would require. This is not a claim that the system eliminates human review; it
is a claim, supported by the coverage figures reported in Section~\ref{sec:results}, that it
reduces the volume of review required from the whole of a match to the specific fraction of it the
system cannot itself resolve with confidence, and that this fraction shrinks measurably as footage
quality improves along the dimensions described in Section~\ref{sec:setting}.

\section{Discussion}
\label{sec:discussion}

\subsection{The same fix, four times}

Read stage by stage, Section~\ref{sec:results} can look like four unrelated results. Read
together, they are one result repeated. Figure~\ref{fig:journey} puts all four on a single axis:
outcome determination, team assignment, and the segmentation boundary problem each start near
chance when approached as something to be perceived, and each lands close to ceiling once the
system instead observes the specific structural fact the sport's own rules already fix, who is
serving, where that server stands, whether a boundary landed inside a rule-verified rally. Identity
is the exception, and it is the exception for a specific, mechanical reason rather than a matter of
degree: volleyball's rules fix who serves next and which side a server stands on, but they do not
fix which of six players on a side is the one who happens to touch the ball, so there is no
equivalent structural fact for identity to observe its way to. Our own accuracy trajectory on
identity (Figure~\ref{fig:trainingladder}) climbs from a zero-shot baseline through a fine-tuning
attempt that briefly made things worse (Figure~\ref{fig:lorafailure}) to the best trained-specialist
result, and even that best result is the lowest of the four right-hand points in
Figure~\ref{fig:journey}, well short of the near-ceiling accuracy the other three stages reach.
That gap is not a shortfall of effort; it is the visible signature of a stage with no rule to fall
back on.

\subsection{What transfers across sports, and what does not}

Volleyball sits at a particular point on the axes described in Section~\ref{sec:related}: six
players a side, a single ball, and a rulebook that constrains the sequence of legal touches tightly
enough that team-level state is almost entirely recoverable without perception at all, as
Section~\ref{sec:results} demonstrated directly. Not every sport shares that last property to the
same degree, and the difference predicts, in advance of running any experiment, where the approach
in this paper will and will not transfer cleanly. Sports whose scoring is built from discrete,
rule-governed events (volleyball, baseball, and to a lesser extent American football) hand a
system a large amount of free structure: who is due to act next, what constitutes a legal sequence,
and how a play must end are all questions the rules answer independent of what the camera observed.
Sports whose value instead accrues continuously, through spacing, movement, and possession that is
not cleanly discretised into individual scoring events (basketball across a full possession,
football played across ninety minutes of open play) hand a system correspondingly less free
structure per unit of game time, and our own supporting measurements on basketball, reported
separately from this paper's primary volleyball results, are consistent with this prediction:
reading a venue's own scoreboard, where one is visible, recovers a team's running point total
essentially exactly, while frame-by-frame event perception on the same footage substantially
overstates it, because a possession that ends in a made basket looks visually similar, at a fixed
camera's resolution, to one that does not. What does not change from sport to sport is the crowding
problem itself: Figure~\ref{fig:overlapbasketball} shows the same pattern of overlapping players and
officials in basketball footage no tracker was ever run against, unrelated to our volleyball corpus
and filmed on a different camera entirely, alongside the volleyball instance in
Figure~\ref{fig:overlapvolleyball}. What a sport's rules change is how much a system needs to resolve
that crowding to get a usable statistic out the other end, not whether the crowding happens at all;
basketball's continuous possession structure means a missed identity in the middle of a crowded
frame is rarely load-bearing the way a missed kill attribution is in volleyball, which is a
statement about rule density, not about basketball footage being any less crowded to look at. The
broader claim this comparison supports is not that
volleyball is uniquely tractable, but that a sport's own rule density is itself a measurable,
predictive property, and that the framework in this paper, decompose the task into stages, ask how
much of each stage the rules already determine, and spend perceptual effort only on the remainder,
generalises across that axis even where the specific numbers reported in Section~\ref{sec:results}
do not.

\subsection{The single-actor-to-field-sport axis, revisited}

Section~\ref{sec:related} introduced, as a narrative frame, the progression from a single golfer
through paired opponents to full team sports and finally to large-roster field sports, and argued
that identity attribution degrades sharply and does not recover as the number of candidate actors
grows. Our own results add a second, orthogonal axis to that progression that the related-work
survey alone could not establish, because it required controlled measurement rather than a survey
of published claims: viewing angle. A camera positioned behind the end line of a court shows the
flat face of a jersey, legible in principle at almost any distance the camera can resolve a player
at all; a camera positioned along the sideline shows the same jersey edge-on, foreshortened by
perspective regardless of resolution, and Section~\ref{sec:setting} showed this difference
outweighing resolution itself in determining which of our held-out matches were tractable. This
matters specifically for the transition from paired to team sports the related-work section
describes, because a two-player sport's court geometry typically constrains camera placement toward
exactly the angle that keeps both competitors legible, while a six- or eleven-player sport's larger
playing surface does not offer that constraint for free; a camera good enough to keep the whole
field in frame is frequently, for purely geometric reasons, positioned at an angle that makes
individual identity harder to resolve than the same camera's resolution alone would suggest. Amateur
venues compound this further in ways broadcast venues systematically do not share: a single fixed
camera with no operator to reposition it, no second camera to cover the angle the first one
misses, and no budget for the elevated, professionally surveyed camera positions broadcast venues
routinely use. The gap this paper measures between amateur and broadcast conditions is therefore not
one gap but at least three compounding ones (resolution, viewing angle, and the absence of a second
vantage point to fall back on when the first one is occluded), and any claim that a method
validated on broadcast footage will transfer to amateur venues needs to account for all three, not
resolution alone.

\subsection{What this is not: a review or judging system}
\label{ssec:notjudging}

Given how easily the two are conflated in public discussion of sports video AI, this paper's task
is best stated by what it is not. Nothing in this work is designed, evaluated, or
intended to adjudicate a contested call, officiate a match in progress, or serve as a review system
of the kind used to overturn or confirm a referee's decision. That is a categorically different
task from the one this paper studies, along at least three axes. It is a real-time task, made under
time pressure during live play, rather than a retrospective one produced after the fact and open to
correction. It is an adversarial and consequential task, where the output directly determines a
match's result and where an error is, by construction, disputed by the party it disadvantages,
rather than a descriptive task whose output a coach can accept, edit, or discard without changing
what already happened on the court. And it demands a standard of certainty appropriate to a binding
decision, not the tiered, openly-uncertain output this paper argues is the honest way to present a
statistic a model cannot fully resolve. The published evidence on exactly this adjacent task
supports treating it as separate and, at present, unready: benchmarks built specifically to test
whether current multimodal systems can make refereeing-grade judgements report the best available
models correct only a little above half the time on hard decisions
\citep{xu2026refereebench}, and our own outcome-recognition results in Section~\ref{sec:results}
tell the same story from the perspective of this paper's own corpus: a frontier model asked to
determine how a rally ended, from the same footage this paper evaluates identity on, defaults to the
most common outcome far more often than it is correct, which is a far more serious error to make
when the judgement in question is binding rather than advisory. Every design choice in this paper
that reads as caution, abstaining rather than guessing, routing uncertain attributions to a human
reviewer rather than resolving them automatically, reporting three tiers of confidence rather than
one number, is a design appropriate to a statistics system whose worst failure mode is an
overlooked correction, not to a judging system whose worst failure mode is a wrong decision that
stands.

\subsection{Why amateur sport needs this to be real work, not a demonstration}

We return, to close, to the scale established in Section~\ref{sec:intro}, because it is the
reason this problem is worth the negative results reported throughout this paper rather than only
the positive ones. A professional or broadcast team that lacks automated statistics has an
alternative: a paid analyst, a video coordinator, a coaching staff whose job includes exactly this
kind of review. The great majority of the eight million young athletes counted in
Section~\ref{sec:intro} have no such alternative. For them, the realistic comparison is not between
an imperfect automated system and a skilled human analyst; it is between an imperfect automated
system, honestly presented with its uncertainty made visible, and no statistics at all, or a
volunteer parent's handwritten tally kept inconsistently from game to game. Framed this way, a
system that resolves a majority of a match's contacts with high confidence, flags the remainder for
a coach's attention rather than guessing at them, and costs a fraction of what a single installed
camera system requires before any subscription is added, is not a weaker substitute for
professional-grade analysis; it is the first time this specific population of athletes has had
access to any statistical account of their own play at all. This is also why we regard the negative
results in this paper, kills that remain coach-confirmed, sets that remain largely unresolved, an
identity ceiling no paradigm we tested crosses, as reportable contributions rather than
embarrassments to be minimised. A coach deciding whether to trust a number needs to know precisely
which numbers to trust, and a paper that reported only what worked would leave that coach worse
equipped to use the system than one that reports, plainly, where it does not yet work at all.

\subsection{What a commercially useful account of this problem can and cannot disclose}

A paper written by authors who operate a commercial system in this space carries an obligation the
purely academic literature reviewed in Section~\ref{sec:related} does not: to be honest about what
is, and is not, being disclosed, rather than presenting a partial account as though it were
complete. This paper discloses every accuracy result, every cost figure, every negative result, and
the shape of the architecture used to obtain them, which stages are handled by which class of
method, and why. It does not disclose the specific thresholds, prompts, model configurations, or
calibration choices that make a deployed version of this system commercially operable, for the same
reason a substantial body of prior work in this exact position has not: describing what a system
achieves and under what conditions is a scientific contribution; describing precisely how to
reproduce its calibrated internals is not required to support that contribution, and doing so would
hand a direct commercial disadvantage to the authors for no corresponding gain to a reader's
understanding of the underlying question. Prior work operating under materially the same
constraint (evaluated on data its authors could not release, built by authors employed by the
organisation whose product the paper evaluates) has taken exactly this position and stated it
plainly rather than obscuring it \citep{wang2024tacticai}. We follow the same posture here: this
paper's contribution is the measured comparison across paradigms and the account of why identity
resists every one of them, not a specification a competitor could implement directly from the text.

\section{Limitations}
\label{sec:limitations}

We state the limitations of this study explicitly, in the same spirit as the null results reported
throughout Section~\ref{sec:results}, rather than allowing them to surface only on close reading.

The corpus, while substantially larger than any comparable amateur sports dataset we are aware of,
is drawn from a single sport, played predominantly by youth and school teams in one geographic
region, filmed on a limited set of camera setups. The stage-by-stage framework in
Section~\ref{sec:stages} and the discussion of what transfers across sports in
Section~\ref{sec:discussion} are offered as a generalisable account, but the specific numbers
reported in Section~\ref{sec:results} are measured on volleyball specifically and should not be
read as sport-agnostic constants.

Kill attribution, discussed at length in Section~\ref{sec:results}, remains unresolved at the
individual-player level in the current system, and we present it as coach-confirmed rather than
fully automated for this reason; a reader should not infer from any team-level total reported in
this paper that per-player kill statistics are production-ready. Set attribution is similarly weak
and is reported as such rather than shipped.

Several of our per-match results, particularly those broken out by camera angle and player height
in Section~\ref{sec:setting}, are computed over a small number of held-out matches, since matches
with independently verified per-touch ground truth are expensive to produce and few exist at this
granularity anywhere in the field. We report these results with explicit per-match breakdowns
rather than a single pooled figure specifically so a reader can judge their statistical weight
directly rather than trusting an aggregate that a handful of matches could dominate.

Our paradigm comparison held inputs constant wherever possible, but neither extreme result is a
hard ceiling. The prompting and agentic figures are bound by the sampling and context limits of
the commercial APIs available during the study period, and a differently engineered prompting
system could plausibly do better. The trained-specialist figures are bound by the labelled data we
have, large for this task but not unlimited, and a larger model fit to more data than currently
exists at this scale could plausibly do better still.

Finally, this paper evaluates methods, not commercial products: it does not benchmark accuracy
against any named vendor, because no vendor publishes the per-touch, per-player ground truth this
methodology requires. Section~\ref{sec:related}'s vendor quotes establish what conditions the
market's products require, not what accuracy they achieve, and the two are not the same claim.

\section*{Statements}
\label{sec:statements}
\addcontentsline{toc}{section}{Statements}

\paragraph{Ethics.}
The footage underlying this study depicts identifiable minors and young adults participating in
organised school and club sport. Every video was recorded as part of the ordinary operation of
matches that had already been scheduled and filmed by teams, coaches, or families for their own
review and record-keeping purposes, under licence terms that permit its use for the development and
evaluation of statistical analysis of the kind reported here; no footage was collected specifically
for this study, and no additional filming, staging, or intervention with any athlete occurred. We
did not seek a formal institutional review determination for this analysis, on the assessment that
it constitutes secondary analysis of pre-existing, licensed recreational-sport recordings rather
than human-subjects research in the sense that determination process is designed to govern; we
state this explicitly, together with the reasoning behind it, rather than asserting review was
unnecessary without explanation. No athlete's full name, or any information that would allow a
reader of this paper to identify a specific individual from a specific match, is published anywhere
in this paper; player identity within our results is referenced only by jersey number, and every
figure derived from raw video frames has had all faces, in the tracked players and in any bystander
visible in the frame, obscured beyond recovery, together with any signage or scoreboard content that
could identify the venue, before the figure was produced. We considered, and are explicit about, a
downstream risk this line of work carries that is distinct from privacy: publishing reliable
per-player statistics about youth athletes could plausibly be put to uses (recruitment screening,
premature specialisation pressure, or comparison across children at an age where statistical
performance is a poor predictor of anything durable) that the athletes and families whose footage
informed this study did not anticipate when the footage was recorded, and we regard this as a
genuine open question for the field rather than one this paper resolves.

\paragraph{Data availability.}
The primary corpus used in this study (66 matches and their associated per-touch annotations) was
obtained under a commercial licence that does not permit redistribution, and it is not released with
this paper. This position mirrors published precedent for sports-analytics research conducted under
comparable data agreements, where the underlying data and, in some cases, the associated code are
likewise withheld while the paper's findings and methodology are reported in full
\citep{wang2024tacticai}. We release, in place of the raw corpus, a full description of its
composition and construction (Section~\ref{sec:setting}), our complete evaluation protocol
(Section~\ref{sec:results}), and every quantitative result, including every negative result, needed
to assess and build on the paper's claims without access to the underlying video.

\paragraph{Code availability.}
The specific system configuration used to obtain our production deployment results in
Section~\ref{sec:deployment} is not released, as it constitutes the operating configuration of a
commercial product. We describe every component's role, training data source, and evaluation
protocol in sufficient detail to support independent replication of the paper's comparative findings
using non-proprietary tooling, consistent with disclosure practice in comparable prior work
conducted under similar constraints.

\paragraph{Competing interests.}
All authors are employed in the research and development function of the company whose commercial
system is evaluated in Section~\ref{sec:deployment} and whose data licence provided the corpus
described in Section~\ref{sec:setting}. This is a competing interest, not merely a funding
disclosure, and we state it as such rather than describing the paper as free of competing interests:
the authors have a financial stake in the perceived capability of the system evaluated here. We
have sought to mitigate this by reporting negative and null results throughout
Section~\ref{sec:results} at the same level of detail as positive ones, by declining to compare our
system's accuracy against any named competitor's, and by subjecting our own production system to the
identical evaluation protocol applied to every other paradigm in this paper.

\paragraph{Use of AI tools.}
Large language models were used as one of the four paradigms under direct experimental evaluation in
this paper, as described in Section~\ref{sec:paradigms}, and results involving them are reported as
data rather than as writing assistance. Language models were separately used during the preparation
of this manuscript to assist with editing and with locating and verifying candidate literature for
citation; every citation appearing in this paper was independently checked by the authors against
its original source before inclusion, and any AI-assisted text was reviewed and edited by the
authors, who take full responsibility for the paper's content and claims.

\paragraph{Funding.}
This work was internally funded by Sports Vision, Inc. No external funding was received.

\paragraph{Acknowledgements.}
We thank the coaches, athletes, and families at Cristo Rey Jesuit High School Twin Cities for
match footage, feedback on early system output, and the kind of practical correction, telling us
where an attribution was wrong and why, that shaped the tiered review design described in
Section~\ref{sec:deployment}.

\bibliographystyle{unsrtnat}
\bibliography{references}

\end{document}